\documentclass[final,5p,times,twocolumn]{elsarticle} 

\makeatletter
\newif\if@restonecol
\makeatother

\usepackage[linesnumbered,ruled]{algorithm2e}
\usepackage{algorithmic}

\usepackage{makecell}
\usepackage{amssymb}
\usepackage{graphicx}
\usepackage{subfigure}
\usepackage{extarrows}
\usepackage{colortbl}
\usepackage{booktabs}
\definecolor{mygray1}{gray}{.8}
\definecolor{mygray2}{gray}{.9}
\definecolor{mygray3}{gray}{.95}
\definecolor{mygray}{gray}{.9}
\definecolor{mypink}{rgb}{.99,.91,.95}
\definecolor{mycyan}{cmyk}{.3,0,0,0}
\usepackage{caption}
\usepackage{url}
\usepackage{array}
\newcommand{\PreserveBackslash}[1]{\let\temp=\\#1\let\\=\temp}
\newcolumntype{C}[1]{>{\PreserveBackslash\centering}p{#1}}
\newcolumntype{L}[1]{>{\PreserveBackslash\raggedleft}p{#1}}
\newcolumntype{R}[1]{>{\PreserveBackslash\raggedright}p{#1}}
\providecommand{\shortcite}[1]{\cite{#1}}

\usepackage{amssymb}

\usepackage[table]{xcolor}
\usepackage{multirow}
\usepackage{amsthm}
\usepackage{graphicx}
\usepackage{extarrows}
\usepackage{wasysym}
\usepackage{placeins}
\usepackage{hyperref}
\usepackage{diagbox}
\DeclareMathOperator{\E}{\mathbb{E}}

\journal{Journal of Web Semantics}

\begin{document}
	
	\begin{frontmatter}
		
		
		
		\title{Less is More: Data-Efficient Complex Question Answering over Knowledge Bases}
		\author[rvt0,rvt3]{Yuncheng Hua}
		\ead{devinhua@seu.edu.cn}
		\author[rvt1]{Yuan-Fang Li}
		\ead{yuanfang.li@monash.edu}
		\author[rvt0,rvt2]{Guilin Qi\corref{cor1}}
		\ead{gqi@seu.edu.cn}
		\author[rvt0]{Wei Wu}
		\ead{wuwei@seu.edu.cn}
		\author[rvt0]{Jingyao Zhang}
		\ead{zjyao@seu.edu.cn}
		\author[rvt0]{Daiqing Qi}
		\ead{daiqing_qi@seu.edu.cn}
		\cortext[cor1]{Corresponding author}
		\address[rvt0]{School of Computer Science and Engineering, Southeast University, Nanjing, China}
		\address[rvt1]{Faculty of Information Technology, Monash University, Melbourne, Australia}
		\address[rvt2]{Key Laboratory of Computer Network and Information Integration (Southeast University), Ministry of Education, Nanjing, China}
		\address[rvt3]{Southeast University-Monash University Joint Research Institute, Suzhou, China}
		
%

\begin{abstract}  
Question answering is an effective method for obtaining information from knowledge bases (KB).
In this paper, we propose the Neural-Symbolic Complex Question Answering (NS-CQA) model, a data-efficient reinforcement learning framework for complex question answering by using only a modest number of training samples. 
Our framework consists of a neural \emph{generator} and a symbolic \emph{executor} that, respectively, transforms a natural-language question into a sequence of primitive actions, and executes them over the knowledge base to compute the answer. 
We carefully formulate a set of primitive symbolic actions that allows us to not only simplify our neural network design but also accelerate model convergence. 
To reduce search space, we employ the copy and masking mechanisms in our encoder-decoder architecture to drastically reduce the decoder output vocabulary and improve model generalizability. 
We equip our model with a memory buffer that stores high-reward promising programs.
Besides, we propose an adaptive reward function.
By comparing the generated trial with the trials stored in the memory buffer, we derive the curriculum-guided reward bonus, i.e., the proximity and the novelty.
To mitigate the sparse reward problem, we combine the adaptive reward and the reward bonus, reshaping the sparse reward into dense feedback.
Also, we encourage the model to generate new trials to avoid imitating the spurious trials while making the model remember the past high-reward trials to improve data efficiency.
Our NS-CQA model is evaluated on two datasets: CQA, a recent large-scale complex question answering dataset, and WebQuestionsSP, a multi-hop question answering dataset. 
On both datasets, our model outperforms the state-of-the-art models. Notably, on CQA, NS-CQA performs well on questions with higher complexity, while only using approximately 1\% of the total training samples.
\end{abstract}  

\end{frontmatter}

\section{Introduction} 
\label{Intro}
%
Knowledge base question answering (KBQA)~\cite{berant2013semantic, yao2014information, yih2014semantic, bordes2015large, yih2015semantic} is an active research area that has attracted significant attention. 
KBQA aims at interpreting natural-language questions as logical forms, action sequences, or programs, which could be directly executed on a knowledge base (KB) to yield the answers. 

Many techniques have been proposed for answering single-hop or multi-hop questions over a knowledge base. 
Neural network-based methods~\cite{berant2013semantic, yao2014information, yih2014semantic, bordes2015large, yih2015semantic} represent the state-of-the-art in KBQA. 
More recently, complex knowledge base question answering (CQA)~\cite{saha2018complex} has been proposed as a more challenging task. 
Complex question answering, the subject of this paper, focuses on aggregation and multi-hop questions, in which a sequence of discrete operations -- e.g., set conjunction, counting, comparison, intersection, and union -- needs to be executed to derive the answer. 

\begin{figure}[htb]
\centering
\includegraphics[width=1.0\columnwidth]{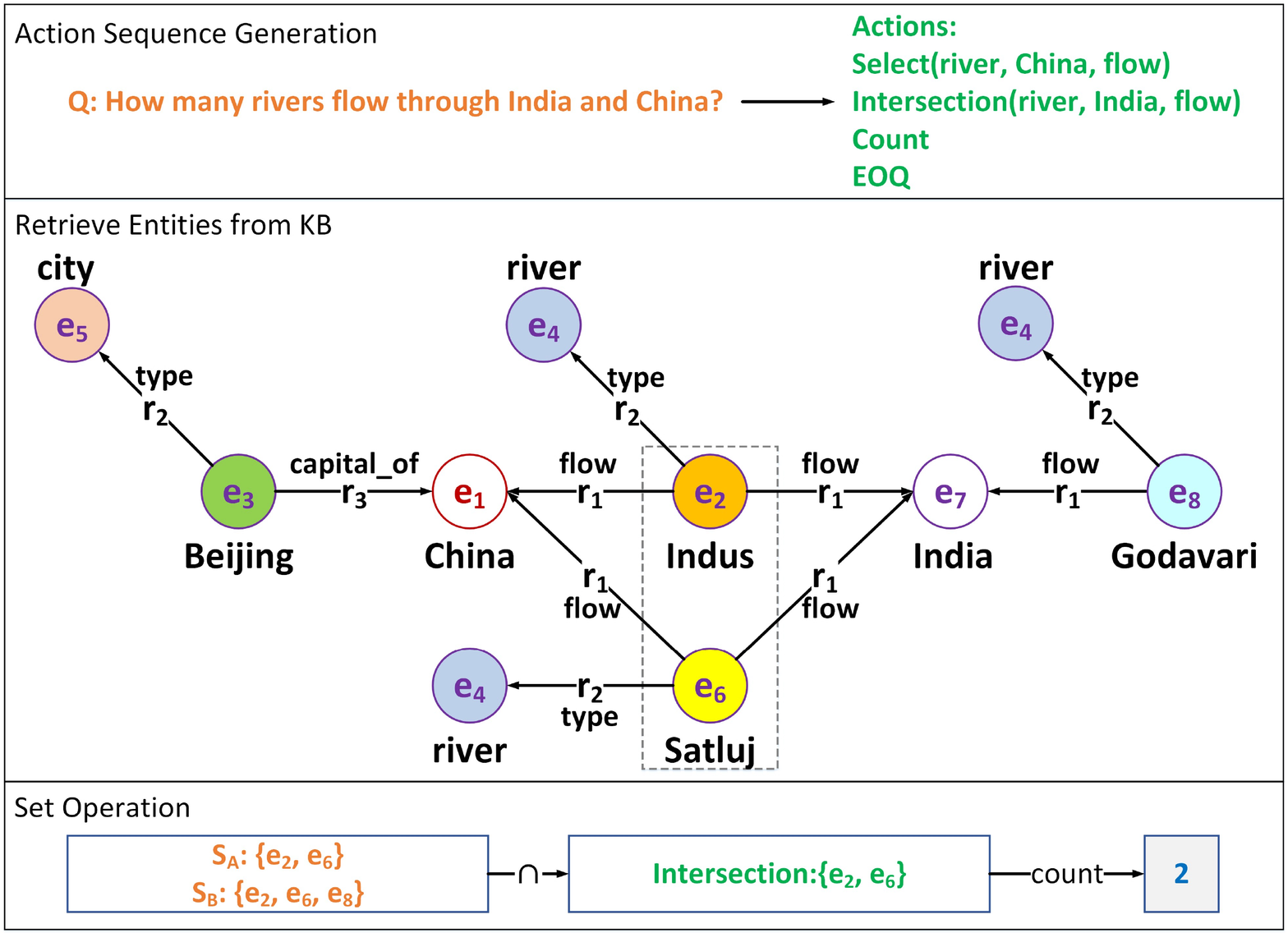}
\caption{An example illustrating the task of complex question answering.} \label{fig1}
\end{figure} 


Complex question answering is typically cast as a \emph{semantic parsing} problem, whereby natural-language questions are transformed into appropriate structural queries (sequences of discrete actions). 
Such queries are then executed on the knowledge base to compute the answer. 
Consider the complex question ``How many rivers flow through India \textbf{and} China?'' as a motivating example. 
Fig.~\ref{fig1} shows an incomplete sub-graph relevant to this question. 
To answer this question, all entities whose type is ``river'' and link to the entity ``China'' with edge ``flow'' will first need to be retrieved from the KB to form the candidate set $S_A$. 
Meanwhile, the candidate set $S_B$ will also be formed to represent those rivers that flow through India. 
After obtaining the intersection of $S_A$ and $S_B$, the number of elements in the intersection can finally be identified as the correct answer to the question. 
It can be seen that a diverse set of operations, including selection, intersection, and counting operations need to be sequentially predicted and executed on the KB. 

\emph{Sequence-to-sequence} (seq2seq) models learn to map natural language utterances to executable programs, and are thus good model choices for the complex question answering task. 
However, under the supervised training setup, such models require substantial amounts of annotations, i.e., manually annotated programs, to effectively train. 
For practical KBQA applications, gold annotated programs are expensive to obtain, thus most of the complex questions are not paired with the annotations~\cite{saha2019complex}. 
Reinforcement learning (RL) is an effective method for training KBQA models~\cite{liang2017neural,ansari2019neural} as it does not require annotations, but only denotatinons (i.e.\ answers) as weak supervision signals. 
However, RL-based KBQA methods face a number of significant challenges. 

\begin{figure*}[htb]
\centering
\includegraphics[width=1.0\textwidth]{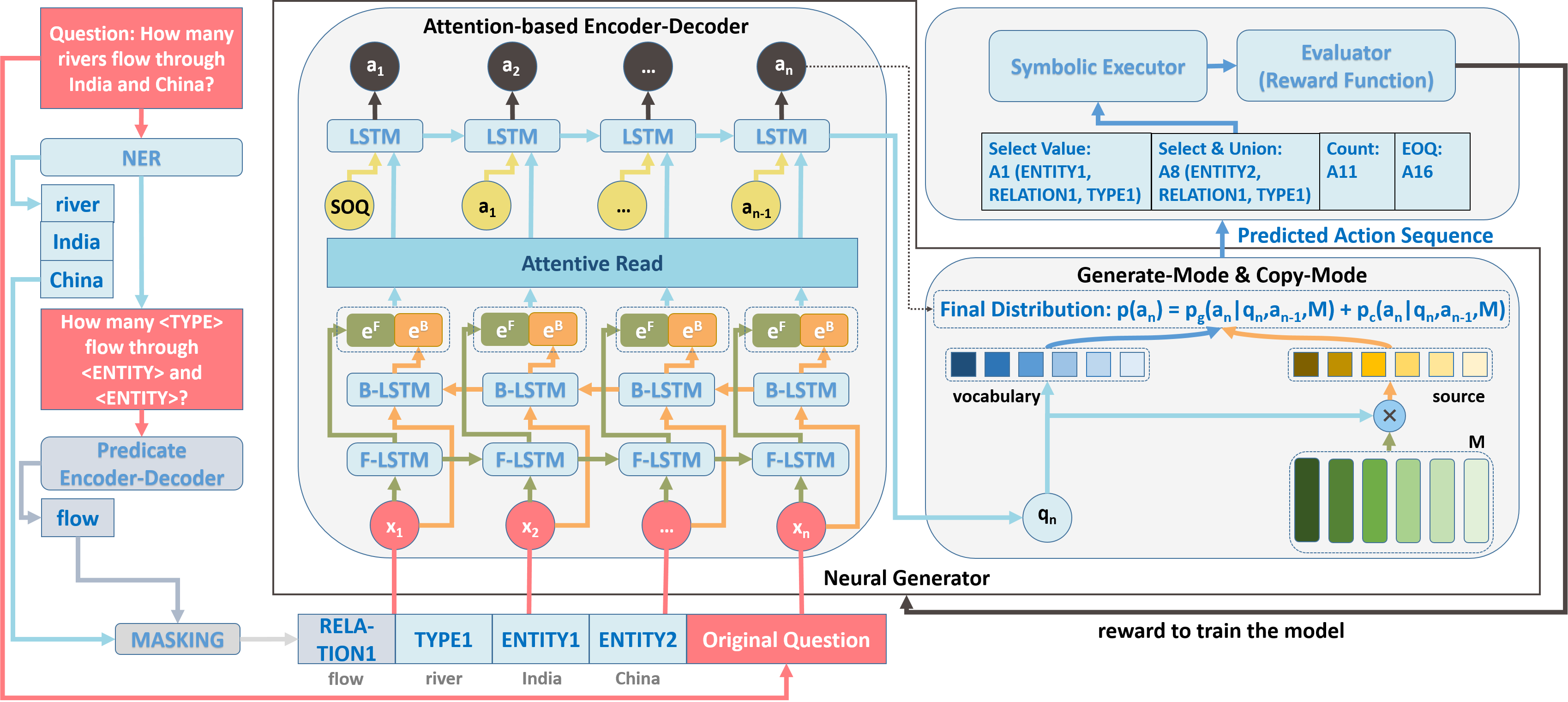}
    \caption{Overview of the proposed NS-CQA model.} \label{fig2}
\end{figure*}


\textbf{Sparse reward.} Neural-symbolic models that are optimized through RL have been proposed for the complex question answering task~\cite{liang2017neural,saha2019complex,ansari2019neural}. 
In the RL context, questions of the same pattern could be regarded as one single \emph{task}, while programs trying to solve these similar questions are considered \emph{trials}. 
Instead of using the gold annotations, neural-symbolic models employ \emph{rewards}, i.e., comparisons between the predicted answer and the ground-truth answer as the distant supervision signal to train the policy~\cite{liang2017neural,ansari2019neural}.
Usually, a positive reward could only be given at the end of a long sequence of correct actions.
However, in the initial stage of model training, most of the trials sampled from the sub-optimal policy attain small or zero rewards~\cite{DBLP:conf/iclr/SavinovRVMPLG19}.
We randomly selected 100 questions from a complex question answering dataset, and manually inspect the generated trials.
We found that more than 96.5\% of the generated trials led to wrong answers and got zero reward.
Thus, this \textbf{sparse reward} problem in the complex question answering task is a major challenge that current neural-symbolic models face. 

Complex Imperative Program Induction from Terminal Rewards (CIPITR)~\cite{saha2019complex} is the state-of-the-art method for the complex question answering task.
It employs high-level constraints and additional auxiliary rewards to alleviate the sparse reward problem.
By using pre-defined high-level constraints, CIPITR restricts the model to search for possible actions that are semantically correct.
Besides, it rewards the model by the additional feedback when the generated answers have the same type of ground-truth answers.
However, on the one hand, defining the high-level constraints comes at the cost of manual analysis to guide the model's decoding process, which is tedious and expensive.
On the other hand, CIPITR only harnesses the type of predicted answers as the auxiliary reward, which makes many failed experience still only have a zero reward. 
At the initial stage of training, most of the programs generated by CIPITR fail to yield expected answers and gain zero rewards, thus most of the programs do not contribute to model optimization. 
Thus, the sparse reward problem remains a challenge for CIPITR.

\textbf{Data inefficiency.} Furthermore, due to the sparse supervision signals, such models are often \textbf{data inefficient}, which means many trials are required to solve a particular task~\cite{liang2018memory}.
Being trained from scratch, RL models often need thousands of trials to learn a simple task, no matter what policy is employed to search programs. 
When faced with a large number of questions, training such models would consume an enormous amount of time. 
One way to increase data efficiency is to acquire task-related prior knowledge to constrain the search space. 
However, in most cases, such prior knowledge is unavailable unless manual labeling is employed. 
Hence, the data inefficiency problem often makes models expensive to train and thus infeasible/impractical.
Liang et al~\shortcite{liang2017neural} proposed the Neural Symbolic Machine (NSM) that maintains and replays \emph{one} pseudo-gold trial that yields the highest reward for each training sample.
When using RL to optimize the policy, NSM assigns a deterministic probability to the best trial found so far to improve the training data efficiency.
However, in NSM, the best trial to be replayed might be a spurious program, i.e., an incorrect program that happens to output the correct answer. 
Under such circumstances, NSM would be misguided by the spurious programs since such programs could not be generalized to other questions of the same pattern.  
Besides, NSM only harnesses the accuracy of the predicted answers to measure the reward, hence also suffers from the sparse reward problem. 


\textbf{Large search space.} 
Large KBs, like Wikidata~\cite{vrandevcic2014wikidata} or Freebase~\cite{bollacker2008freebase}, contain comprehensive knowledge and are suitable to be sources for complex question answering.
The KBs with smaller size usually embrace a limited number of facts, thus are insufficient to yield the required answers.
To answer the questions, searching for KB artifacts (entities, classes, and predicates in KB) that are related to the questions is a crucial step. 
However, large KBs often lead to vast search space.
Given a sequence of actions, at each step of its execution, both the operator and the parameters, i.e., KB artifact, used in action need to be correct.
To produce the correct result, the actions in the sequence also need to follow a particular order. 
With the above factors combined, searching desired action sequences would consume a considerable amount of time and memory, which makes the training process slow and expensive. 
How to design a set of simple yet effective actions that could reduce the search space remains a challenge.

In this paper, we propose a Neural-Symbolic Complex Question Answering framework, NS-CQA.
It trains a policy to generate the desired programs from the comparison between the generated answer and the ground-truth result.
We incorporate a memory buffer, design an adaptive reward function, and propose a curriculum-guided reward bonus to improve the model. 


To \textbf{reduce search space}, we use a combination of simple yet effective techniques to reduce model complexity, increase generalizability, and expedite training convergence.
We employ the widely-used masking technique to allow the model to handle unseen KB artifacts effectively. 
In conjunction, the copy mechanism~\cite{gu2016incorporating} is also employed to reduce the size of the decoder output vocabulary drastically.
Accordingly, the decoder output vocabulary only needs to contain the primitive actions and a handful of masks, instead of the entire KB. 
Also, we carefully craft a set of primitive actions that are necessary for the complex question answering task and simplify the query form to reduce the search space. 
Our primitive actions free the model from the need for maintaining expensive and sophisticated memory modules. 
On the contrary, the model maintains the intermediate results in a simple key-value dictionary. 
The model can directly compute the final answer to a question by executing a correct sequence of primitive actions.

Different from previous neural-symbolic models for the complex question answering task, our NS-CQA framework is designed to \textbf{augment the sparse extrinsic reward} by a dense intrinsic reward and exploit trials efficiently.

Instead of using the manually defined high-level constraints as in CIPITR, we employ a random search algorithm to find a few pseudo-gold annotations that lead to correct answers.
The pseudo-gold annotations are used as supervision signals to pre-train the model, which could guide the model to filter out infeasible action sequences. 
Consequently, in the initial training stage, we mitigate the cold start issue by using the pseudo-gold annotations as demonstration data to pre-train our model.

Moreover, instead of recording one trial for each question, we maintain a memory buffer to store the promising trials, i.e., the action sequences that lead to the correct answer or gain high rewards.
In our work, on the one hand, we aim to converge a behavior policy to a target optimal policy. 
Thus we need to measure how similar/important the generated trials are to trials that the target policy may have made. 
We design a reward bonus, \textbf{proximity}, to favor these ``similar/important'' trials.
On the other hand, to avoid overfitting the spurious trials, we also encourage the policy to explore in undiscovered search space, i.e., the space that is beyond the imitation of the pseudo-gold annotations.
Therefore, we design another reward bonus, \textbf{novelty}, to the generated action sequences that are ``different'' from the trials stored in the memory buffer.

To adaptively control the exploration-exploitation trade-off, we employ a \emph{curriculum learning}~\cite{bengio2009curriculum} scheme to dynamically change the influence of the two reward bonuses, namely the proximity and the novelty. 
Particularly, given a question, we define the proximity for the predicted trial as the highest similarity between the predicted trial and all the trials in the memory buffer.
Novelty is also defined through similarity. 
We consider a trial is novel if it is not similar to all the trials in the memory buffer. 
At the initial stage, since the policy is sub-optimal and the trials generated by the policy are generally infeasible, we prefer more novelty for exploration. 
We gradually increase the proportion of proximity and reduce the influence of novelty during the later training epochs. 
At the final stage of the training, the proportion of proximity in the reward bonus will rise to 100\%. 
Such a delicately designed curriculum method can significantly improve the learning quality and efficiency~\cite{fang2019curriculum}.

Besides, to alleviate the sparse reward problem, an adaptive reward function (ARF) is proposed.
ARF encourages the model with partial reward and adapts the reward computing to different question types.
We sum up the reward bonus with the adaptive reward to make our RL model learn from the combined reward. 
With this modification, we reshape the sparse rewards into dense rewards and enable any failed experience to have a nonnegative reward, thus alleviate the sparse reward problem.

Furthermore, by incorporating a memory buffer, which maintains off-policy samples, into the policy gradient framework, we \textbf{improve the sample efficiency} of the REINFORCE algorithm in our work~\cite{liang2018memory}. 
We encourage the model to generate trials that are similar to the trials in the memory buffer by using the proximity bonus. 
Therefore the high-reward trials could be re-sampled frequently to avoid being forgotten in the training process. 
Since a group of question shares the same pattern, a sequence of the same actions could solve such questions. 
Once we improve sample efficiency, we reduce the trials needed for training. 
Therefore, using a minimal subset of training samples, our model can produce competitive results. 


Overall, the main contributions in this paper can be summarized as follows. 
\begin{enumerate}
    \item A neural-symbolic approach that is augmented by memory buffer and is designed for complex question answering. In our method, for each question, we resort to the previous promising trials stored in the memory to assist our model in replaying and generating feasible action sequences. 
    
    \item A curriculum-learning scheme that adaptively combines novelty and proximity to balance the exploration-exploitation trade-off for the model. We treat the combination of the novelty and proximity as a bonus to the reward, therefore alleviate the sparse reward and data inefficiency problems.
    
    \item Several simple yet effective techniques are proposed to reduce search space, improve model generalizability, and accelerate convergence.
    In our work, the masking method and the copy mechanism are incorporated in Seq2Seq learning to avoid searching over a large action space.
    Also, a set of primitive actions is carefully designed to solve complex questions by executing actions sequentially, avoiding the maintenance of complex memory modules.
\end{enumerate}

Our experiments on a large complex question answering dataset~\cite{saha2018complex} and a relatively smaller multi-hop dataset WebQuestionsSP~\cite{yih2016value} show that NS-CQA outperforms all the recent, state-of-the-art models. 
Moreover, NS-CQA performs well on the questions with higher complexity, demonstrating the effectiveness of our method.

The rest of this paper is organized as follows.
Related works are introduced in Section~\ref{sec:related}.
Our NS-CQA framework is described in Section~\ref{sec:model}. 
Section~\ref{sec:experiments} describes the experiments and evaluation results. 
In Section~\ref{sec:qual_ana}, we perform some qualitative analysis, showing positive examples and typical errors. 
We conclude our work in Section~\ref{sec:conclusion}.

\section{Related work}
\label{sec:related}
%
The NS-CQA model is inspired by two lines of work: semantic parsing and neural-symbolic systems. 
Semantic parsing mainly focuses on reformulating natural language questions into logic forms, which are then executed on knowledge bases (KBs) to compute answers~\cite{lehmann2012deqa, bao2014knowledge, hu2018state}.  
More recent approaches employ sophisticated deep learning models to search entities and predicates that are most relevant to the question~\cite{dong2015question, he2016character, lukovnikov2017neural}. 

Many of these works tackle simple one-hop questions that are answerable by a single triple. 
Others address the multi-hop task, in which answers are entities that can be retrieved by a path of connected triples. 
In both cases, a model only retrieves entities as answers in a fixed search space, i.e., the KB. 
Our model addresses a more challenging problem, where answers come from a much larger search space, involving not only entities in the KB but also their logical combinations and aggregations (numbers).

The complex KBQA dataset, such as CQA~\cite{saha2018complex}, requires the execution of discrete actions rather than merely searching for entities in KB.
Memory network~\cite{bordes2015large, kumar2016ask, miller2016key, xu2019enhancing} is used to store facts in KB and makes it possible to solve complex questions. 
\citet{luo2018knowledge} encodes complex questions into vectors to represent the semantics and structure of the input sentence contemporaneously, by which the similarity between the question and the graph could be computed.
Dialog-to-Action (D2A)~\cite{guo2018dialog} incorporates dialog memory management in generating logical forms that would be executed on a large KBs to answer complex questions.
It labels all training samples with pseudo-gold actions and trains the model by imitation learning.
It is worth noting that D2A aims to answer \emph{context-dependent} questions, where each question is part of a multiple-round conversation.
On the other hand, in this paper, we consider answering the single-turn questions.
Therefore we do not include D2A as a baseline method in the evaluation as it is not directly comparable to our problem setup.
The semantic parsing approaches mentioned so far all require the annotation of the entire training dataset to learn the models.
Different from these approaches, our model can be learned from a small number of pseudo-gold annotations only.

Some recent studies in KBQA focus on semi-supervised learning, in which models are trained solely on denotations, e.g., the execution results of queries, the state of the final time step, etc. 
In~\cite{berant2013semantic}, a semantic parser is trained by learning from question-answer pairs rather than annotated logical forms to query the knowledge graph. 
Furthermore, the parser could be trained without manual annotations or question-answer pairs by treating denotations of natural language questions and related KB queries as weak-supervision~\cite{reddy2014large}. 
Similarly, queries automatically generated from knowledge graph triples and paraphrased questions without answers are used in weak-supervision to train subgraph embedding models~\cite{bordes2014open1}. 
To fully understand the intention of questions, STAGG~\cite{yih2015semantic} is proposed to generate a staged query graph and directly map the graph into $\lambda$-calculus.

Neural-symbolic models integrate neural networks with logic-based symbolic executors to conduct non-differentiable computations. 
Neural Turing Machines (NTMs)~\cite{graves2014neural} pioneer the neural-symbolic methods, in which REINFORCE is employed to train the model in the RL Neural Turing Machines (RL-NTMs)~\cite{zaremba2015reinforcement}. 
When answering natural language questions on relational tables, some approaches~\cite{neelakantan2015neural, pasupat2015compositional} predict discrete symbolic operations by neural networks and obtain answers by executing them. 
\citet{guu2017language} marry RL and maximum marginal likelihood (MML) to avert the spurious problems. 
The neural-symbolic visual question answering (NS-VQA) system~\cite{yi2018neural} combines deep representation learning and symbolic program execution to solve visual question answering problems over a synthetic image dataset.

Most relevant to this work are two state-of-the-art techniques on complex KBQA:  Neural Symbolic Machines (NSM)~\cite{liang2017neural} and CIPITR~\cite{doi:10.1162/tacl_a_00262}.
NSM deals with multi-hop questions in the \textsc{WebQuestionsSP} dataset~\cite{yih2016value} with two components: the programmer and the computer. 
By employing an EM-like mechanism, NSM iteratively finds the pseudo-gold trials for the training questions.
NSM then assigns the pseudo-gold trials with a deterministic probability, therefore, to anchor the model to the high-reward trials.
In a similar vein, CIPITR translates a natural-language complex question into a multi-step executable program using the Neural Program Induction (NPI) technique. 
CIPITR does not require gold annotations and can learn from auxiliary rewards, KB schema, and inferred answer types.

Our neural-symbolic model is different from them in that we augment our RL-based model with a memory buffer to record the successful trials.
With the help of the memory buffer, we could compute the extra reward bonus to encourage the model to generate new trials, imitate successful trials, and reshape the sparse reward to provide dense feedback.
As can be seen in Table~\ref{tab2}, our NS-CQA model outperforms all the baseline models, and the performance difference is prominent on the more complex categories of questions. 
Also, in Table~\ref{tab4}, we can find that NS-CQA is better than other baseline models.
The superiority of NS-CQA in both the datasets verified the effectiveness and generalization ability of the model.

\section{NS-CQA: A Complex Question Answering Approach}
\label{sec:model}
In our work, the complex question answering problem is regarded as a semantic parsing task: given a complex question $q$ consisting of tokens $(w_1,\ldots,w_{m})$, the model generates a primitive action sequence $(a_1,\ldots,a_{q})$, and execute the sequence on the KB $\mathcal{K}$ to yield the answer $a$.

This section outlines our NS-CQA approach to the complex question answering problem. 
We first describe the set of primitive actions in Section~\ref{sec:actions}. 
Given a complex question, the \emph{semantic parser} (Section~\ref{ssec:semantic parser}) recognizes KB artifacts that are relevant to the question. 
By combining the question and the output of the parser, the \emph{neural generator} (Section~\ref{ssec:neural generator}) transforms the query into a sequence of primitive actions.

The \emph{symbolic executor} (Section~\ref{ssec:symbolic executor}) executes the actions on the KB to obtain an answer. 
Overall, we employ \emph{RL} to directly optimize the generator through a policy gradient on the answer predicted by the executor (Section~\ref{ssec:training paradigm}).
The high-level architecture of our model is depicted in Fig.~\ref{fig2}.



\textbf{}
\textbf{}

\begin{table*}[htb]
\centering
\caption{Demonstration of how intermediate result evolves when executing actions sequentially.}
\label{tab:dictionary}
\begin{tabular}{|p{2cm}|p{1.8cm}|p{3.5cm}|p{3.5cm}|p{3.5cm}|}
\hline 
\multirow[c]{4}{=}{\textbf{Question1:} Which country has \textbf{maximum number} of rivers?} & \multirow[t]{2}{=}{\textbf{Actions}} & \multirow[t]{2}{=}{\textbf{Action1:}} & \multirow[t]{2}{=}{\textbf{Action2:}} & \multirow[t]{2}{=}{\textbf{Action3:}}\\ 
&   & \itshape SelectAll (country, flow, river) & \itshape ArgMax & \itshape EOQ\\ \cline{2-5}
& \multirow[t]{4}{=}{\textbf {Retrieved Key-Value Pairs}} & \{China:\{Indus, Satluj\}\} & \multirow[t]{4}{=}{/} & \multirow[t]{4}{=}{/} \\ 
&   & \{India:\{Indus, Satluj, Godavari\}\} &  & \\ 
&   & \{Russia:\{Volga, Moskva, Neva, Ob\}\} &  & \\ 
&   & \{USA:\{Mississippi, Colorado, Rio Grande\}\} &  & \\ \cline{2-5}
& \multirow[t]{4}{=}{\textbf {Dictionary}} & \{China:\{Indus, Satluj\}\} & \multirow[t]{4}{=}{\{Russia:\{Volga, Moskva, Neva, Ob\}\}} & \multirow[t]{4}{=}{\{Russia:\{Volga, Moskva, Neva, Ob\}\}} \\ 
&   & \{India:\{Indus, Satluj, Godavari\}\} &  & \\ 
&   & \{Russia:\{Volga, Moskva, Neva,Ob\}\} &  & \\ 
&   & \{USA:\{Mississippi, Colorado, Rio Grande\}\} &  & \\ \hline
\multirow[c]{4}{=}{\textbf{Question2:} What rivers flow in India \textbf{but not} China?} & \multirow[t]{2}{=}{\textbf{Actions}} & \multirow[t]{2}{=}{\textbf{Action1:}} & \multirow[t]{2}{=}{\textbf{Action2:}} & \multirow[t]{2}{=}{\textbf{Action3:}}\\ 
&   & \itshape Select (India, flow, river) & \itshape Diff (China, flow, river) & \itshape EOQ\\ \cline{2-5}
& \textbf {Retrieved Key-Value Pairs} & \{India:\{Indus, Satluj, Godavari\}\} & \{China:\{Indus, Satluj\}\} & / \\ \cline{2-5}
& \textbf {Dictionary} &\{India:\{Indus, Satluj, Godavari\}\}& \{India:\{Godavari\}\} & \{India:\{Godavari\}\} \\ \hline
\end{tabular}
\end{table*}

\subsection{Primitive Actions}\label{sec:actions}

We propose a set of primitive actions based on the subset of SPARQL queries that are necessary for the current complex question answering task and simplify the query form to reduce the search space.
Our actions are designed to be simple, dispensing with SPARQL features, including namespaces, etc.
It also does not support certain SPARQL features, including OPTIONAL, FILTER, etc.
Since our actions belong to a subset of SPARQL's operators, the complexity of our actions follows that of SPARQL.
The main contributions of this paper relate to the neural generator of these programs. 
Thus we leave the study of operator complexity to future research.

Unlike NSM~\cite{liang2017neural}, which introduces the variables to save the intermediate results, we employ a \textbf{key-value memory} to maintain the intermediate result.
NSM adds a new intermediate variable into the decoder vocabulary after an action is executed, thus enables the decoder to generate the new variable in later decoding steps.
Consequently, NSM dynamically increases the size of the decoder vocabulary in the decoding process.
On the contrary, with the help of the memory, our model does not need to refer to previous intermediate variables, thus fix the size of the decoder vocabulary to simplify the model. 

We use two components, i.e., an operator and a list of variables, to compose the primitive actions. 
After analyzing different types of complex problems, we design 17 operators in this work, which are described in Table~\ref{table:primitive_actions}.
Besides, the key-value dictionary $\mathcal{D}$ is designed to store intermediate results.
Keys in $\mathcal{D}$ refer to entities present in the question or specific special symbol, and the values are the obtained elements related to it. 
Before the execution of the whole action sequence, $\mathcal{D}$ is initialized as empty.
When executing an action, the model will generate an intermediate result based on content in $\mathcal{D}$, which is the result derived from the last action.
Then the generated intermediate result will be further stored in $\mathcal{D}$ to update it.
The contents of updated $\mathcal{D}$ are then preserved for later use.

For instance, when dealing with the question ``Which country has \emph{maximum number} of rivers?", the desired output actions should be ``{\itshape SelectAll(country, flow, river), ArgMax, EOQ}".
The first action has an operator \emph{SelectAll}, a relation variable \emph{flow} and two type variables, i.e., \emph{country} and \emph{river}, while no entity variable is found in this action.
Note that the second action only has one operator \emph{ArgMax}, and so does the third action \emph{EOQ}.
By performing the first action, we retrieve KB to find all entities belong to type `country' as keys.
Meanwhile, we set the river-type entities linked with country-type entities by relation `flow' as values.
Like what is demonstrated in Table~\ref{tab:dictionary} (the retrieved results presented in the table are not consistent with the actual KB while are only for demonstration), one country-type entity is \emph{USA} where the linked river-type objects are \emph{\{Mississippi, Colorado, Rio Grande\}}.
The retrieved key-value pairs (the key is one country-type entity and value is a set of river-type entities) are then stored in dictionary $\mathcal{D}$ as the intermediate result of this action.
After that, the second action is executed to find the key whose mapped value has most elements.
In could be found in Table~\ref{tab:dictionary} \emph{Russia} has most elements thus \emph{Russia:\{Volga, Moskva, Neva, Ob\}} is then kept in $\mathcal{D}$ and other key-value pairs are removed.
Then we update $\mathcal{D}$ and view this key-value pair as the intermediate result of the second action.
When encountered with \emph{EOQ}, the model outputs the final result in $\mathcal{D}$.

To simplify the action sequence, we design particular actions (GreaterThan, LessThan, Inter, Union, and Diff) as relatively `\textbf{high-level}' actions that need multiple set operations to perform. 
Though such design will enhance the difficulty of symbolic executor, on the other hand, it could reduce the complexity of the neural generator.
Since the bottleneck of our model lies in the difficulty of training generator, we make a compromise between executor and generator. 

Take the question ``What rivers flow in India \emph{but not} China?'' as an example. 
The reference action sequence should be ``{\itshape Select(India, flow, river), Diff(China, flow, river), EOQ}''. 
After executing the first action, a set of rivers flowing in India is stored in $\mathcal{D}$ as the value of the key \emph{India}.
As showed in Table~\ref{tab:dictionary}, such key-value pair is \emph{India:\{Indus, Satluj, Godavari\}}. 
When executing the second action ``\emph{Diff(China, flow, river)}'', all river entities linked to \emph{China} by relation \emph{flow} are first retrieved from the KB. 
Then based on the key-value pair stored in $\mathcal{D}$, the entities indexed to India but not China will be kept as the updated value of the key \emph{India}.
In this case is \emph{India:\{Godavari\}}. 
Then the key-value pair in $\mathcal{D}$ is updated accordingly.
Upon encountering the action \emph{EOQ}, the key-value pair stored in $\mathcal{D}$ is returned as the final answer to this question.

As described above, the model executes all the actions in sequence.
With the help of a key-value dictionary $\mathcal{D}$, the intermediate result of the current action is recorded and preserved for later use.
The model could perform the following actions based on the result stored in $\mathcal{D}$, and the new result would further update $\mathcal{D}$.
Therefore, the design of $\mathcal{D}$ makes executing actions sequentially possible.

\begin{table*}[!htb]
\centering
    \caption{The set of primitive actions. $\mathcal{K}$ represents the knowledge base, $e, e_1, e_2, \ldots$ represent entities, $r$ represents a relation, $t$ represents a type, and $\mathcal{D}$ represents a key-value dictionary which stores intermediate results.}\label{table:primitive_actions}
    \begin{tabular}{|l|l|p{10cm}|p{3.5cm}|}
    \hline
    {\bfseries ID} & \textbf{Action} & {\bfseries Retrieved Key-Value Pairs} & {\bfseries Output} \\
    \hline
    A1 & $Select(e, r, t)$ & $ \{e_2|e_2 \in t, (e,r,e_2)\in\mathcal{K}\}$ & $\mathcal{D} = \mathcal{D} \cup \{e:\{e_2\}\}$\\
    \hline
    A2 & $SelectAll(et, r, t)$ & $ \{e_2|e_1 \in et, e_2 \in t, (e_1,r,e_2)\in\mathcal{K}\}$ & $\mathcal{D} = \mathcal{D} \cup \{e_1:\{e_2\}\}$ \\
    \hline
    A3 & $Bool(e)$ & $ value = 1\,if\,e \in \mathcal{D}\,;otherwise\,value = 0$ & $\mathcal{D} = \{bool: value\}$ \\
    \hline
    A4 & $ArgMin$ & $\{e_1|e_1 \in \mathcal{D}, \exists (e_1:\{e_2\}) \in \mathcal{D},\forall e':(e':\{e'_2\}) \in \mathcal{D}, |\{e_2\}| \leq |\{e'_2\}|\}$ & $\mathcal{D} = \{e_1:\{e_2\}\}$ \\
    \hline
    A5 & $ArgMax$ & $\{e_1|e_1 \in \mathcal{D}, \exists (e_1:\{e_2\}) \in \mathcal{D},\forall e':(e':\{e'_2\}) \in \mathcal{D}, |\{e_2\}| \geq |\{e'_2\}|\}$ & $\mathcal{D} = \{e_1:\{e_2\}\}$ \\
    \hline
    A6 & $GreaterThan(e)$ & $\{e_1|e_1 \in \mathcal{D}, \exists (e_1:\{e_2\}) \in \mathcal{D},\exists e'_2:(e:\{e'_2\}) \in \mathcal{D}, |\{e_2\}| \geq |\{e'_2\}|\}$ & $\mathcal{D} = \{e_1:\{e_2\}\}$ \\
    \hline
    A7 & $LessThan(e)$ & $\{e_1|e_1 \in \mathcal{D}, \exists (e_1:\{e_2\}) \in \mathcal{D},\exists e'_2:(e:\{e'_2\}) \in \mathcal{D}, |\{e_2\}| \leq |\{e'_2\}|\}$ & $\mathcal{D} = \{e_1:\{e_2\}\}$ \\
    \hline
    A8 & $Inter(e,r,t)$ & $ \{e_2|e_2 \in t, (e,r,e_2)\in\mathcal{K}\}$ & $\mathcal{D} = \mathcal{D} \cap \{e:\{e_2\}\}$ \\
    \hline
    A9 & $Union(e,r,t)$ & $ \{e_2|e_2 \in t, (e,r,e_2)\in\mathcal{K}\}$ & $\mathcal{D} = \mathcal{D} \cup \{e:\{e_2\}\}$ \\
    \hline
    A10 & $Diff(e,r,t)$ & $ \{e_2|e_2 \in t, (e,r,e_2)\in\mathcal{K}\}$ & $\mathcal{D} = \mathcal{D}-\{e:\{e_2\}\}$ \\
    \hline
    A11 & $ Count$ & $ Card(\mathcal{D}) = |\{e|e \in \mathcal{D},\exists (e:\{e_2\}) \in \mathcal{D}\}|$ & $\mathcal{D} = \{num: Card(\mathcal{D})\}$ \\
    \hline
    A12 & $AtLeast(n)$ & $\{e|e \in \mathcal{D}, \exists (e:\{e_2\}) \in \mathcal{D}, |\{e_2\}| \geq n\}$ & $\mathcal{D} = \{e:\{e_2\}\}$ \\
    \hline
    A13 & $AtMost(n)$ & $\{e|e \in \mathcal{D}, \exists (e:\{e_2\}) \in \mathcal{D}, |\{e_2\}| \leq n\}$ & $\mathcal{D} = \{e:\{e_2\}\}$ \\
    \hline
    A14 & $EqualsTo(n)$ & $\{e|e \in \mathcal{D}, \exists (e:\{e_2\}) \in \mathcal{D}, |\{e_2\}| = n\}$ & $\mathcal{D} = \{e:\{e_2\}\}$ \\
    \hline
    A15 & $GetKeys$ & $ \{e|e \in \mathcal{D}, \exists (e:\{e_2\}) \in \mathcal{D}\}$ & $\mathcal{D} = \{key: \{e\}\}$ \\
    \hline
    A16 & $Almost(n)$ & $\{e|e \in \mathcal{D}, \exists (e:\{e_2\}) \in \mathcal{D}, ||\{e_2\}|-n| \leqslant \alpha\} $ where $\alpha$ is predefined & $\mathcal{D} = \{e:\{e_2\}\}$ \\
    \hline
    A17 & {\itshape EOQ} & \emph{end of sequence} & $\mathcal{D}$ \\
    \hline
  \end{tabular}
  \end{table*}

\subsection{Semantic Parser}\label{ssec:semantic parser}
Given a natural language question, the parser first recognizes entity mentions (for example $India$) and class mentions (for example $river$)~\cite{huang2015bidirectional}. 
A \textbf{Bidirectional-LSTM-CRF} model is employed to label the entity/type mentions~\cite{yin2016simple}.
The parser then links them with the corresponding entities and types in KB.
At first, the parser tries to retrieve the entity/type candidates related to mentions by computing the literal similarities. 
Besides, the description of the candidates and the question are embedded into vectors to get semantic similarity.
The literal and semantic similarities are thus integrated to rank the entity/type candidates, while the ones have the highest score is selected as linked entities/types.
Subsequently, the entity/class mentions in the query are replaced with wild-card characters to generate patterns.
We employ a \textbf{convolutional Seq2Seq model}~\cite{gehring2017convolutional} to transform the the generated patterns to corresponding relations.

\subsection{Neural Generator}\label{ssec:neural generator}
Our generator is an attention-based Seq2Seq model augmented with the copy and masking mechanisms. 
Given a question with tokens $(w_{1},\ldots,w_m)$, the generator predicts tokens $(a_{1},\ldots,a_{q})$. 
The input of the model is the original complex question concatenated with KB artifacts generated by the semantic parser, and the output is the tokens of a sequence of actions. 
In our work, the output at each time step is a single token which is used to compose actions with adjacent output tokens.

For a vanilla Seq2Seq model, all the KB artifacts corresponding to all questions will need to be collected in advance to make the vocabulary large enough to cover all questions.
Let $N$ represent the maximum number of actions in sequences. Since we have designed 17 different operators in our work, and each of operators can take up to three arguments (see Table~\ref{table:primitive_actions} for details), in the worst case, the vocabulary size of the decoder is $O(17^N*|\mathcal{E}|^{2N}*|\mathcal{P}|^N)$, where $|\mathcal{E}|$ and $|\mathcal{P}|$ denote the number of entities (including types) and predicates in the KB $\mathcal{K}$ respectively. Given a large KB such as Freebase, $|\mathcal{E}|$ and $|\mathcal{P}|$ can be very large. Such a vocabulary size would be prohibitively large for the decoder, making it highly unlikely to generate the correct token, thus negatively affecting the rate of convergence. 

By incorporating the masking mechanism, the names of KB artifacts used for compose actions are replaced with \emph{masks} such as \texttt{<ENTITY1>}, \texttt{<TYPE1>} and \texttt{<PREDICATE1>}. 
Thus, an action sequence consisted of $N$ actions will be masked into the following form: $A^{(1)}(\texttt{<E$_1$>}, \texttt{<P$_1$>}, \texttt{<E$_2$>}),\ldots, \allowbreak A^{(N)}(\texttt{<E$_{2N-1}$>}, \texttt{<P$_N$>}, \texttt{<E$_{2N}$>})$.
For instance, the action sequence `{\itshape Select(India, flow, river), Diff(China, flow, river), EOQ}' is used to solve the problem 'What rivers flow in India but not China?'.
After masking, the real names of artifacts in actions are substituted with \emph{masks}, and the action sequence is changed into `{\itshape Select(ENTITY1, PREDICATE1, TYPE1), Diff(ENTITY2, PREDICATE1, TYPE1), EOQ}'. 
The mappings between the actual names and \emph{masks} are recorded in our model, which will be later used to recover the exact names of KB artifacts in actions when being executed.
Given the maximum number of actions $N$, with the masking mechanism, the decoder vocabulary size is reduced to $O(17^N*(2N)^{2N}*N^N)$, where $(2N)\ll|\mathcal{E}|$ and $N\ll|\mathcal{P}|$.
As action sequences are typically not long (i.e., \ $N\leq 5$ in our observation), this represents orders of magnitude reduction in vocabulary size.

Also, we could alleviate the Out Of Vocabulary (OOV) problem with the help of the masking mechanism. 
OOV words refer to unknown KB artifacts that appear in the testing questions but not included in the output vocabulary and would make the generated action incomplete. 
When facing a question with unseen KB artifacts, the model is not able to predict such objects since they are out of the output vocabulary.
However, when employing the masking mechanism, all the KB artifacts are translated into masks, thus enabling the model to select masks from relatively fixed output vocabulary.
Like the above example presented, the KB artifacts `India' and `China' are replaced with the mask `ENTITY1' and `ENTITY2'.
Thus our model only needs to generate the masked tokens instead of the real names of the KB artifacts, which will mitigate the OOV problem.
In consequence, the model could form the actions more precisely.

 
To further decrease search space, the copy mechanism is also incorporated.
The copy mechanism \emph{replicates} all masked symbols in the input sequence to form the output, instead of letting the decoder generate them from the decoder vocabulary. 
As a result, the decoder only needs to generate primitive actions, further reducing vocabulary size to $O(17^N)$. 

The benefits of our design are twofold. 
(1) The much-reduced vocabulary makes convergence faster, as the generator is only concerned about generating correct primitive actions, but not names of artifacts from the KB. 
(2) Solve questions with unforeseen KB artifacts by directly masking and copying them from the input question when generating an action sequence. 

\paragraph{\textbf{Encoder}}
The encoder is a bidirectional LSTM that takes a question of variable length as input and generates an encoder vector $\boldsymbol{e}_{i}$ at each time step $i$.
\begin{equation}
\begin{aligned}
    \boldsymbol{e}_{i},\boldsymbol{h}_{i} = LSTM(\phi_{E}(x_{i}),\boldsymbol{h}_{i-1}).
\end{aligned}
\end{equation}
Here $\phi_{E}$ is word embedding of token $E$. 
$(\boldsymbol{e}_{i},\boldsymbol{h}_{i})$ is the output and hidden vector of the $i$-th time step when encoding. 
The dimension of $\boldsymbol{e}_{i}$ and $\boldsymbol{h}_{i}$ are set as the same in this work.  
$\boldsymbol{e}_{i}$ is the concatenation of the forward ($\boldsymbol{e}_{i}^{F}$) and backward ($\boldsymbol{e}_{i}^{B}$) output vector and $\boldsymbol{h}_{i}$ is the encoder hidden vector. 
The output vectors $(\boldsymbol{e}_{1},\ldots,\boldsymbol{e}_{T})$ is regarded as a short-term memory $\boldsymbol{M}$, which is saved to use in copy mode. 

\paragraph{\textbf{Decoder}}
Our decoder of NS-CQA predicts output tokens following a mixed probability of two models, namely \textbf{generate-mode} and \textbf{copy-mode}, where the former generates tokens from the fixed output vocabulary and the latter copies words from the input tokens. 
Furthermore, when updating the hidden state at time step $t$, in addition to the word embedding of predicted word at time $t-1$, the location-based attention information is also utilised.

By incorporating the copy mechanism, the generated actions might be chosen from the output vocabulary or input tokens. We assume a fixed output vocabulary $\mathcal{V}_{output} = \{v_1,\ldots,v_N\}$, where $\mathcal{V}_{output}$ contains operators and related arguments in action sequence. In addition to that, all the unique words from input tokens $x = \{x_{1},\ldots,x_{T}\}$ constitute another set $\mathcal{X}$ whereby some OOV words could be `copied' when such words are contained in $\mathcal{X}$ but not in $\mathcal{V}_{output}$. Therefore, the vocabulary unique to input tokens $x$ is: $\mathcal{V}_{x} = \mathcal{V}_{output} \cup \mathcal{X}$. 

\indent The $\textbf{generate-mode}$ predicts output token $a_{t}$ from the output vocabulary $\mathcal{V}_{output}$. Like traditional Seq2Seq model, the decoder is another LSTM model that generates a hidden vector $\boldsymbol{q}_{t}$ from the previous output token $a_{t-1}$. Previous step's hidden vector $\boldsymbol{q}_{t-1}$ is fed to an attention layer to obtain a context vector $\boldsymbol{c}_{t}$ as a weighted sum of the encoded states. Current step's $\boldsymbol{q}_{t}$ is generated via:
\begin{equation}
\begin{aligned}
    \boldsymbol{q}_{t} = LSTM(\boldsymbol{q}_{t-1}, 
    [\phi_{D}(a_{t-1}),\boldsymbol{c}_{t}])
\end{aligned}
\end{equation}
Here $\phi_{D}$ is the word embedding of input token $a_{t-1}$. The dimension of $\boldsymbol{q}_{t}$ is set as $d_q$, and the attention weight matrix is trainable. 
The hidden vector $\boldsymbol{q}_{t}$ is used to compute the score of target word $v_i$ in $\mathcal{V}_{output}$ as: 
\begin{equation}
    \psi_{g}(a_t = v_i) = \boldsymbol{v}_i^{\top} \boldsymbol{W}_{o}\boldsymbol{q}_{t},\  v_i\in\mathcal{V}_{output}
\end{equation}  
where $\boldsymbol{W}_{o}$ is trainable matrix and $\boldsymbol{v}_i$ is the vector of word $v_i$. 

In $\textbf{copy-mode}$, the score of ``copying'' word $x_j$ from input tokens $\{x_{1},\ldots,x_{T}\}$ is computed as:   
\begin{equation}
    \psi_{c}(a_t = x_j) = \sigma(\boldsymbol{e}_{j}\boldsymbol{W}_{c}\boldsymbol{q}_{t}),\  x_j\in\{x_{1},\ldots,x_{T}\}
\end{equation} 
where $\boldsymbol{W}_{c} \in \mathbb{R} ^{d_q\times d_q}$, $\sigma$ is a non-linear activation function and, hidden encoder vectors $\{\boldsymbol{e}_{1},\ldots,\boldsymbol{e}_{T}\}$ in short-term memory $\boldsymbol{M}$ are used to map the input tokens $\{x_{1},\ldots,x_{T}\}$ respectively.

Finally, given the hidden vector $\boldsymbol{q}_t$ at time $t$ and short-term memory $\boldsymbol{M}$,  the output token  $a_t$ is generated following a mixed probability as follows:
\begin{equation}
\begin{aligned}
    p(a_t|\boldsymbol{q}_t,a_{t-1},\boldsymbol{M}) = p_{cg}(a_t|\boldsymbol{q}_t,a_{t-1},\boldsymbol{M})
\end{aligned}
\end{equation} 
where $p_g$ and $p_c$ indicate the generate-mode and copy-mode respectively. They are calculated as follows:
\begin{equation}
    p_{cg} = 
    \left\{
        \begin{array}{lcl}
        \frac{1}{Z}e^{\psi_{g}(a_t)},       &      & {a_t\in \mathcal{V}_{output}}\\
        \frac{1}{Z}{\Sigma_{j:x_j = a_t}} e^{\psi_{c}(a_t)},       &      & {a_t\in \mathcal{X}}\\
        \end{array} \right. 
\end{equation} 
where $Z$ is the normalization term and is computed as: $Z = \Sigma_{v\in\mathcal{V}_{output}}e^{\psi_{g}(v)} + \Sigma_{x\in\mathcal{X}}e^{\psi_{c}(x)}$.

\indent At time step $t$, word embedding of previous output token $a_{t-1}$ and the location-based attention information are both employed to update the hidden state. 
$a_{t-1}$ will be represented as $[\phi_{D}(a_{t-1});\boldsymbol{r}_{q_{t-1}}]$, where $\phi_{D}(a_{t-1})$ is the word embedding of $a_{t-1}$ and $\boldsymbol{r}_{q_{t-1}}$ is the weighted sum of hidden states $\{\boldsymbol{e}_{1},\ldots,\boldsymbol{e}_{T}\}$ in $\boldsymbol{M}$.

Vector $\boldsymbol{r}_{q_{t-1}}$ is calculated as:
\begin{align}
    \boldsymbol{r}_{q_{t-1}} &= \Sigma_{\tau=1}^{T}\rho_{t\tau}\boldsymbol{e}_{\tau} \\
    \rho_{t\tau} &= 
    \left\{
        \begin{array}{lcl}
        \frac{1}{K}p_{cg}(x_{\tau}|\boldsymbol{q}_{t-1},a_{t-1},\boldsymbol{M}),       &      & {x_{\tau}=a_{t-1}}\\
        0,     &      & otherwise\\
        \end{array} \right. 
\end{align}
where $K$ is the normalization term which is $\Sigma_{{\tau}':x_{{\tau}'}=a_{t-1}}p_{cg}(x_{{\tau}'}|\boldsymbol{q}_{t-1},a_{t-1},\boldsymbol{M})$, considering there might be input tokens located at different positions which equal to $a_{t-1}$.
$\rho_{t\tau}$ is viewed as location-based attention in our work.

\subsection{Symbolic Executor}\label{ssec:symbolic executor}
A symbolic executor is implemented as a collection of deterministic, generic functional modules to execute the primitive actions, which have a one-to-one correspondence with the functional modules. 
The symbolic executor first analyzes the output tokens produced by neural generator, and would assemble the actions one by one.
Given an action sequence that begins with the first action, the symbolic executor executes the actions in order, on the intermediate result of the previous one. 
As discussed in Section~\ref{sec:actions}, this is only possible due to our carefully designed primitive actions. 
Otherwise, complex memory mechanisms would need to be incorporated to maintain intermediate answers~\cite{guo2018dialog, liang2017neural}. 
Upon encountering the action $EOQ$, the result from the last execution step will be returned as the final answer.

\subsection{Training Paradigm}\label{ssec:training paradigm}
As the symbolic executor executes non-differentiable operations against a KB, it is difficult to utilize end-to-end back-propagation to optimize the neural generator. 
Therefore, we adopt the following two-step procedure to train the generator. 
By using a breadth-first-search (BFS) algorithm, we generate pseudo-gold action sequences for a tiny subset of questions.
In BFS, we assemble all the operators and KG artifacts found in question to form candidate action sequences in a brute-force way.
We then execute the candidate action sequences to find the ones that yield the right answer and view them as pseudo-gold action sequences.
Using these pairs of questions and action sequences, we pre-train the model by Teacher Forcing.

We then employ RL to fine-tune the generator on another set of question-answer pairs.
The symbolic executor executes the predicted action sequence to output an answer and yield a reward for RL.
The reward is the similarity between the output answer and the gold answer.

As shown in Algorithm~\ref{alg1}, our method works as follows.
The training starts with an empty memory buffer, and at every epoch, for each sample, the generated trials that gain high reward will be stored in the memory.
For each question, we first use a search algorithm, i.e., greedy-decoding, to generate a trial.
We execute the trial and compute a reward $r_{greedy}$, which is set as the reward threshold.
Then we employ a beam search method to generate a set of candidate trials for the question and compute their rewards.
At every epoch, we compare the generated trials with the trials in memory to determine the reward bonus, aka proximity and novelty, and further add the reward bonus to the adaptive reward.
A candidate trial is added into the memory buffer if its reward is higher than $r_{greedy}$. 
We utilize the augmented reward to train the policy under the RL setting.

The memory buffer stores a limited set of trials for the training questions.
Once the memory buffer is full, we substitute a random trial with the current new trial.
This strategy enables the memory buffer to maintain relatively fresher trials, but would not always abandon the older ones.

\begin{algorithm}[htb]
\caption{Training NS-CQA}
\label{alg1}
\SetAlgoLined
\DontPrintSemicolon
\KwIn{Training dataset $Q_{train}$, initial policy $\theta$, memory buffer $M$, reward function $R(\cdot)$, learning rate $\eta_1$} 
\KwOut{The learned policy $\theta^*$}
Randomly initialize $\theta$\;
$M\leftarrow\emptyset$\;
\While{not converged}{
    Sample batch of data $Q_{batch} \sim Q_{train}$\;
    $\mathcal{L}\leftarrow0$\;
    \For{$q \in Q_{batch}$}
    {
        Get one trial $\boldsymbol{t_{greedy}}$ by greedy-decoding\;
        Compute adaptive reward $r_{greedy}$\;
        Compute cumulative reward $R(q,\boldsymbol{t_{greedy}})$\;
        Sample $K$ trials: $\boldsymbol{t_k} \sim \pi(\boldsymbol{t}|q;\theta)$ \;
        \For{each trial $\boldsymbol{t_k}$}
        {
            Compute adaptive reward $r_{\boldsymbol{t_k}}$\;
            Compute cumulative reward $R(q,\boldsymbol{t_k})$\; 
            Update memory: add $\boldsymbol{t_k}$ to $M$ if $r_{\boldsymbol{t_k}} > r_{greedy}$\;  
        }
        $L=\frac{1}{K}{\sum_{k=1}^K [R(q,\boldsymbol{t_k})-R(q,\boldsymbol{t_{greedy}})]}log(p_\theta(\boldsymbol{t_k}))$\;
        $\mathcal{L}\leftarrow\mathcal{L}+L$\;
    }
    Compute adapted parameters: $\theta\leftarrow \theta+\eta_1 \nabla_{\theta}\mathcal{L}$\;
}
\textbf{Return}  The learned policy $\theta^* \leftarrow \theta$ 
\end{algorithm}

The main components of our training paradigm, namely the RL method, the adaptive reward, and the curriculum reward bonus, are described in the rest of this section. 

\paragraph{\textbf{Reinforcement Learning}}
In this step, REINFORCE~\cite{williams1992simple} is used to finetune the neural generator.
Typically, the three notions mentioned in RL are action, state, and reward, respectively.
In our scenario, at each step, action as a fundamental concept in RL is a token produced by a neural generator used to form an executable action sequence.
What needs to be clarified is that when related to RL, the concept of actions refers to the tokens of a trial.
Since the complex question answering environment is deterministic, we define the state as the question combined with the generated tokens so far.   
Meanwhile, the reward is the same as the notion in RL.   

In our work, the state, action and reward at time step $t$ are denoted as $s_{t}$, $a_{t}$ and $r_{t}$ respectively. 
Given a question ${q}$, the state of time step $t$ is defined by ${q}$ and the action sequence so far: $s_{t} = (q, a_{0:t-1})$, and the action tokens are generated by the generator. 
At the last step of decoding $T$, the entire sequence of actions is generated. 
The symbolic executor will then execute the action sequence to produce an output answer $ans_o$. 
Therefore the reward is computed only after the last step of decoding when $ans_o$ is output.
We design a Adaptive Reward Function (ARF), which is the adaptive comparison of the output answer $ans_o$ and the gold answer $ans_g$. 
Furthermore, we employ a curriculum-guided Reward Bonus (CRB), which comprises proximity and novelty, to assign non-zero rewards for actions that do not yield correct answers. 
Specifically, the cumulative reward of an action sequence $a_{0:T}$ is the sum of CRB and ARF:
\begin{equation}
\label{cumulative_reward}
    R(q,a_{0:T}) = CRB + ARF(ans_o,ans_g)
\end{equation}
Then $R(q,a_{0:T})$ is sent back to update parameters of the neural generator through a REINFORCE objective as the supervision signal.

\paragraph{\textbf{Adaptive Reward}}
Though the search space is significantly reduced to $O(17^N)$ after the masking and copy mechanisms are incorporated, the length of action sequence, which is $N$, would be fairly long when solving a relatively more complex question.
In that case, $O(17^N)$, the size of the search space, is still huge which makes it hard for the model to find correct action sequences when gold annotations are unavailable.

Moreover, since the reward used to train the model could only be obtained after a sequence of actions is executed, the execution of actions is regarded as a part of training.
Suppose a large amount of candidate action sequences (normally more than 50) are generated, their execution would consume a large amount of time since it involves searching triples in KB, performing set operations and discrete reasoning. 
To reduce the training time, we limit our NS-CQA model to form only 5$\sim$20 candidate action sequences with a smaller beam size. 
With the huge search space and small beam size, the sparsity of the reward becomes a problem. 
Of all the candidate action sequences that are predicted, very few of them could output correct answers and be positively rewarded while most of them do not produce any reward. 
Under such circumstances, without the notion of partial reward, the neural generator would suffer from high variance and instability. 
Therefore the generator would be inclined to be trapped in local optima and not generalize well on data never seen before.

Furthermore, different categories of questions entail different answer types. 
The reward function should be adaptive to the diverse answer types which could measure the degree of correctness of predicted answers more precisely. 
In other words, the reward function should encourage the generator to generate action sequences with the correct answer type while punishing the model if the predicted answer type is incorrect. 

In the complex question answering scenario, the possible types of answers are integers, sets of entities and Boolean values. 
To measure the answer correctness more accurately and adaptively, and to compute partial reward to alleviate the sparse reward problem, we define our adaptive reward function $ARF$. 
$ARF$ computes reward based on different answer types using the function $Sim$ between the output answer $ans_o$ and the gold answer $ans_g$.
\begin{align}
    Sim(ans_o,ans_g) &= 
    \left\{
        \begin{array}{lr}
        1 - \frac{|ans_g - ans_o|}{|ans_g + ans_o + \varepsilon|},       & \text{ integer}\\
        Edit(ans_g, ans_o),     &  \text{Boolean}\\
        F1(ans_g, ans_o),     &  \text{ set}\\
        \end{array} \right. 
\end{align}

The edit-score is used to measure the accuracy of the output provided the answer type is Boolean, while the F1-score is used as a reward when answer is a set of entities.
When the answer type is Boolean, the expected output is a list of Boolean value, for instance as what is presented in Table~\ref{tab:cases}, the expected answer of the question ``s Alda Pereira-Lemaitre a citizen of France and Emmelsbull-Horsbull?" is [True, False].
Regard each Boolean value as a single element in a list, the edit (Levenshtein) distance is used to compute the similarity between two lists, i.e., output answer list, and gold answer list.
Thus the similarity is calculated as follows:
\begin{equation}
    Edit(ans_g, ans_o) = 1 - \frac{Levenshtein(ans_g, ans_o)}{max(|ans_g|,|ans_o|)}
\end{equation}
On the other hand, suppose the type of answer is a set of entities, F1-score is computed as:
\begin{equation}
    F1(ans_g, ans_o) = 2 * \frac{precision * recall}{precision + recall} = 2 * \frac{\frac{|ans_g \cap ans_o|}{|ans_o|} * \frac{|ans_g \cap ans_o|}{|ans_g|}}{\frac{|ans_g \cap ans_o|}{|ans_o|} + \frac{|ans_g \cap ans_o|}{|ans_g|}}
\end{equation}
If the answer type is incorrect or the action sequence is sematically invalid, reward is set as 0. 
On the other hand, if the answer type is the same as the gold answer, partial reward is granted. 
We then defined $ARF$ as follows:
\begin{equation}
    ARF(ans_o,ans_g) = R_{type} * (W_1 + W_2 * Sim(ans_o,ans_g))
\end{equation}

In our work, $\varepsilon$, $W_1$ and $W_2$ are predefined hyper-parameters and set as $0.001$, $0.2$ and $0.8$ repectively.
If the type of predicted answer is correct, $R_{type}$ is set as $1$, otherwise $0$.

\paragraph{\textbf{Curriculum-guided Reward Bonus}}
In many RL settings, the reward is positive only when a trial, i.e., a long sequence of actions generated by a policy, could yield the correct result.
At the initial stage, since the policy is not yet fully-trained, out of all the generated trials, the rate of successful trials is rare.
Thus, there is an insufficient number of collected successful trials for training the RL model, which causes the sparse reward problem.

Besides, the suboptimal policy will not explore the search space effectively since many sampled trials could be repeated.
Moreover, the policy will forget the rare successful trials easily since the trials may not be re-sampled frequently.
These factors often lead to the data inefficiency problem.

To solve the above problems, we design two reward bonuses to learn from failed trials. 
We introduce a memory buffer to record the high-reward trials for each training sample.
We compare a generated trial with those stored in the memory buffer to see how similar it is to the recorded trials. 
Therefore we give \emph{proximity bonus} to a generated trial even if it fails to yield the correct answer.
By doing this, we could encourage the policy to re-sample the high-reward trials and accordingly reduce the frequency of generating infeasible trials.
Also, we give \emph{novelty bonus} to the generated trials that differ from the trials in the memory buffer.
The novelty bonus encourages the policy to generate different trials, thus avoid being trapped by spurious ones.
Once the reward is augment with the above two bonuses, the corresponding failed experience is assigned with a nonnegative reward and can contribute to learning the policy.

To balance proximity and novelty, we employ a curriculum-learning method to regulate their trade-off dynamically.
In the earlier epochs, higher novelty can help the policy to explore unseen areas and generate more diverse trials.
However, in the later epochs, such novelty will bring more noise (often as spurious trials) into training and distract the policy.
At the later stage of training, the policy has gained sufficient knowledge about the tasks and is able to generate high-reward, promising trials. 
Therefore, proximity becomes more critical since it will encourage the policy to proceed towards the correct trials and to focus on learning how to generate promising trials. 

Given a question $q$, suppose the high-reward trials ${t^q_1,\ldots,t^q_m}$ have already been stored in the memory buffer $M$.
For one generated trial $t$, we compute the reward bonus CRB as:
\begin{equation}
\label{reward_bonus}
CRB = \alpha(\lambda F_{prox}(t,M)+(1-\lambda)F_{novel}(t,M)),
\end{equation}
where $\alpha \in [0,1]$ is the weight of the reward bonus and dependent on the scale of the task rewards. 
The term $F_{prox}$ reflects the proximity of the trial $t$ to the recorded trials in memory $M$ while $F_{novel}$ measures the novelty of $t$. 
The value of $\lambda$ controls their relative proportion, which is adjusted by the curriculum learning method.

We compute the similarity between the generated trial $t$ with one trial $t^q_i$ in $M$ by edit distance, which is:
\begin{equation}
    s_i = Edit(t,t^q_i)
\end{equation}

Thus we define the proximity $F_{prox}$ as the highest similarity between the trial $t$ and all the trials ${t^q_1,\ldots,t^q_m}$ in the memory buffer, which is:
\begin{equation}
    F_{prox}(t,M) = \mathop{max}\limits_{1\leqslant i\leqslant m}(s_i)
\end{equation}

The term novelty $F_{novel}$ measures the diversity of the trial $t$ from the trials ${t^q_1,\ldots,t^q_m}$.
We assign a high novelty to a generated trial if it is different from the trials in $M$, thus we define the novelty as:
\begin{equation}
    F_{novel}(t,M) = \beta - \frac{1}{m}\Sigma^m_{i=1}s_i,
\end{equation}
where $\beta \in [0,1]$ is used to measure the diversity and is dependent on the scale of the similarity.  

We employ a curriculum learning scheme to adaptively change the weight $\lambda$ in Formula~\ref{reward_bonus}.
We start from learning to generate novel trials with large diversity, and gradually focus on re-sampling the trials which have high proximity to the desired successful trials stored in the memory buffer.
This method is achieved by progressively increasing the weight $\lambda$, which is exponentially increased $\lambda$ with the training epochs: 
\begin{equation}
    \lambda = min\{1,(1+\eta)^{\gamma}\lambda_0\},
\end{equation}
where $\eta \in [0,1]$ is the learning pace which controls the curriculum learning, $\gamma$ represents the number of the epochs that have been trained, and $\lambda_0$ is the initial weight of $\lambda$.

In our work, $\alpha$, $\beta$, $\eta$, and $\lambda_0$ are hyper-parameters which are defined as 0.1, 1.0, 0.08, and 0.1, respectively in our work.

\paragraph{\textbf{REINFORCE}}
At each time step, the output token generated by agent is decided by a certain policy (which is the generator in our work), and the probability that one token $a$ is chosen is computed as below, where $\theta$ denotes model parameters:
\begin{equation}
    \pi_\theta(q,a) = P_{\theta}(a_t=a|q,a_{0:t-1})
\end{equation}

Thus, the probability of an entire action sequence $a_{0:T}$ is given by:
\begin{equation}
    P_{\theta}(a_{0:T}|q) = \displaystyle\prod_{t=1}^T P_{\theta}(a_t|q,a_{0:t-1})
\end{equation}

In \ref{cumulative_reward}, we define the cumulative reward $R(q,a_{0:T})$.  
Our objective is to maximize the expected cumulative reward.
Therefore we use the policy gradient method such as the REINFORCE algorithm to finetune the generator. 
The objective and gradient are:
\begin{equation} \label{gradient}
\begin{aligned}
    J^{RL}(\theta) &= \sum_{q}\E_{P_{\theta}(a_{0:T}|q)}[R(q,a_{0:T})]\\
    \nabla_{\theta}J^{RL}(\theta) &= \sum_{q}\sum_{a_{0:T}}P_\theta(a_{0:T}|q)\cdot[R(q,a_{0:T})\\
    &\quad-B(q)]\cdot\nabla_{\theta}logP_{\theta}(a_{0:T}|q) 
\end{aligned}
\end{equation}
$B(q)=R(q,\hat{a}_{0:T})$ is a baseline that reduces the variance of the gradient estimation without introducing bias. 
In our work, the baseline is set as what is used in the self-critical sequence training (SCST)~\cite{rennie2017self}.
Also, Monte Carlo integration is employed to approximate the expectation over all possible trials in the policy gradient method~\cite{williams1992simple}.
The training method is presented in Algorithm~\ref{alg1}.

\section{Experiments}
\label{sec:experiments}
We evaluated our model NS-CQA on a large-scale complex question answering dataset (CQA)~\cite{saha2018complex}, and a challenging multi-hop question answering dataset WebQuestionsSP~\cite{yih2016value}. 

The CQA dataset is generated from the facts stored in Wikidata~\cite{vrandevcic2014wikidata}, consisting of 944K QA pairs for training and 100K/156K QA pairs for validation and test, respectively. 
The CQA dataset is characterised by the challenging nature of the questions. 
To answer them, discrete aggregation operators such as set union, intersection, min, max, counting, etc.\ are required (see Table~\ref{table:primitive_actions} for more details). 
The CQA questions are organized into seven categories, as shown in Table~\ref{tab2}. 
Some of these categories (e.g., Simple Question) have entities as answers, while others have numbers (e.g., Quantitative (Count)) or Boolean values (e.g., Verification (Boolean)) as answers. 
We used `accuracy' as the evaluation metric for categories whose answer type is `Verification', `Quantitative (Count)', and `Comparative (Count)'; and `F1 measure' for other types of questions. 
However, to simplify the presentation and stay consistent with literature~\cite{saha2019complex,ansari2019neural}, we denote `accuracy' as `F1 measure' in Table~\ref{tab2}. 
Hence, the model performance was evaluated on the F1 measure in this paper.
Furthermore, we computed the micro F1 and macro F1 scores for all the models based on the F1-scores of the seven question categories.

In our analysis of the CQA dataset, we found that the seven categories of questions vary substantially in complexity. 
We found that `Simple' is the simplest that only requires two actions to answer a question, whereas `Logical Reasoning' is more difficult that requires three actions. 
Categories `Verification', `Quantitative Reasoning', and `Comparative Reasoning' are the next in the order of difficulty, which need 3--4 actions to answer. 
The most difficult categories are `Quantitative (Count)' and `Comparative (Count)', needing 4--5 actions to yield an answer. 
Saha et al.~\shortcite{saha2019complex} drew a similar conclusion through manual inspection of these seven question categories. 

The WebQuestionsSP dataset collects multi-hop questions, i.e., the questions require a chain of KB triples to answer, via the Google Suggest API. 
In comparison to the CQA dataset, WebQuestionsSP can be considered easier as it only contains multi-hop questions, and the answers are (sets of) entities only.
It consists of 3,098 question-answer pairs for training and 1,639 questions for testing. 
We utilized the same evaluation metrics employed by~\cite{saha2018complex,liang2017neural}, the F-1 measure, to evaluate model performance on the testing questions. 


\begin{table*}[t!]
\centering
\caption{Performance comparison (measured in F1) of the four methods on the CQA test set. Best results for each category is \textbf{bolded}, and second best is \underline{underlined}.}\label{tab2}
\centering
\begin{tabular}{l*{7}{c}}
\toprule
{\textbf{Method}} & {\textbf{HRED+KVmem}} & {\textbf{CIPITR-All}} & {\textbf{CIPITR-Sep}} & {\textbf{NSM}} &{\textbf{Vanilla}}& {\textbf{PG}} & {\textbf{NS-CQA}}\\ 
\midrule
Simple Question     & 41.40\%   & 41.62\%   & \textbf{94.89\%}  & 88.33\%               & 85.13\%           & 84.25\%           & \underline{88.83\%}\\
Logical Reasoning   & 37.56\%   & 21.31\%   & \textbf{85.33\%}  & 81.20\%               & 70.46\%           & 68.37\%           & \underline{81.23\%}\\
Quantitative Reasoning & 0.89\% & 5.65\%    & 33.27\%           & 41.89\%               & 47.96\%           & \underline{56.06\%}  & \textbf{56.28\%}\\ 
Comparative Reasoning & 1.63\%  & 1.67\%    & 9.60\%            & 64.06\%               & 54.92\%           & \textbf{67.79\%}  & \underline{65.87\%}\\ 
\midrule
Verification (Boolean)& 27.28\% & 30.86\%   & 61.39\%           & 60.38\%               & 75.53\%           & \underline{83.87\%}  & \textbf{84.66\%}\\
Quantitative (Count)  & 17.80\% & 37.23\%   & 48.40\%           & 61.84\%               & 66.81\%           & \underline{75.69\%} & \textbf{76.96\%}\\
Comparative (Count)   & 9.60\%  & 0.36\%    & 0.99\%            & 39.00\%               & 34.25\%           & \underline{43.00\%}  & \textbf{43.25\%}\\ 
\midrule
Overall macro F1      & 19.45\% & 19.82\%   & 47.70\%           & 62.39\%               & 62.15\%         & \underline{68.43\%} & \textbf{71.01\%}\\ 
Overall micro F1      & 31.18\% & 31.52\%   & 73.31\%           & 76.01\%               & 74.14\%         & \underline{76.56\%} & \textbf{80.80\%}\\ 
\bottomrule
\end{tabular}
\end{table*}

\subsection{Model Description}
Our model is evaluated against three baseline models: HRED+KVmem~\cite{saha2018complex}, NSM~\cite{liang2017neural} and CIPITR~\cite{saha2019complex}. 
We used the open-source code of HRED+KVmem and CIPITR to train the models and present the best result we obtained. 
As the code of NSM has not been made available, we re-implemented it and further incorporated the copy and masking techniques we proposed.
HRED+KVmem does not use beam search, while CIPITR, NSM, and our model all do for predicting action sequences.
When inferring the testing samples, we used the top beam~\cite{saha2019complex}, i.e., the predicted program with the highest probability in the beam, to yield the answer. 

\begin{description}
\item[\textbf{HRED+KVmem}~\cite{saha2018complex}] is the baseline model proposed together with the CQA dataset~\cite{saha2018complex}, which combines a hierarchical encoder-decoder with a key-value memory network.
The model first encodes the current sentence with context into a vector, whereby a memory network retrieves the most related memory. 
Then the retrieved memory is decoded to predict an answer from candidate words. 
HRED+KVmem was designed specifically for the CQA dataset, thus was not included in our experiments on WebQuestionsSP. 

\item[\textbf{NSM}~\cite{liang2017neural}] is an encoder-decoder based model which is trained by weak-supervision, i.e., the answers to the questions.
NSM first employs an Expectation-Maximization-like (EM-like) method to find pseudo-gold programs that attain the best reward.
It iteratively uses the current policy to find the best programs and then maximizes the probability of generating such programs to optimize the policy. 
Then NSM replays \emph{one} pseudo-gold trial that yields the highest reward for each training sample when employing REINFORCE to train the policy. 
It assigns a deterministic probability to the best trial found so far to improve the training data efficiency. 
NSM was at first proposed to solve the problems in WebQuestionsSP, and we reimplemented it to also handle the CQA dataset. 

As presented in~\ref{sec:actions}, unlike NSM, we do not refer to the intermediate variables when generating the tokens of a trial. 
Therefore it is unnecessary to incorporate the key-variable memory, which is used to maintain and refer to intermediate program variables in our work. 
We thus removed the key-variable memory component in the seq2seq model in our reimplementation of NSM. 

\item[\textbf{CIPITR}~\cite{saha2019complex}] employs an NPI technique that does not require gold annotations. 
Instead, it relies on auxiliary awards, KB schema, and inferred answer types to train an NPI model. 
CIPITR transforms complex questions into neural programs and outputs the answer by executing them. 
It designs high-level constraints to guide the programmer to produce semantically plausible programs for a question. 
The auxiliary reward is designed to mitigate the extreme reward sparsity and further used to train the CIPITR model. 
CIPITR is designed to handle the KBQA problems proposed in both CQA and WebQuestionsSP.  
\end{description}

\subsection{Training}

The NS-CQA model was implemented in PyTorch with the model parameters randomly initialized\footnote{To encourage reproductivity, we have released the source code at {\small\textsf{\url{https://github.com/DevinJake/NS-CQA}}}.}.
The Adam optimizer is applied to update gradients defined in Formula~\ref{gradient}. 
We used the fixed GloVe~\cite{yin2016simple} word vectors to represent each token in input sequences and set each unique, unseen word a same fixed random vector. 
We set a learning rate of 0.001, a mini-batch size of 32 samples to pre-train the Seq2Seq model with pseudo-gold annotations. 
On average, after about 70 epochs, the Seq2Seq model would converge. 
Then we trained the REINFORCE model with a learning rate of 1e-4 and a mini-batch size of 8 on the pre-trained Seq2Seq model until accuracy on the validation set converged (at around 30 epochs). 
 
As solving the entity linking problem is beyond the scope of this work, we separately trained an entity/class/relation linker. 
When training the NS-CQA model, the predicted entity/class/relation annotations along with the pseudo-gold action sequence (which are generated by a BFS algorithm) were used. 
The entity/class/relation annotations predicted by the respective linker were used when conducting experiments on the test dataset. 

Incorporating the copy and masking mechanisms, our full model took a total of at most 3,700 minutes to train 100 epochs (70 epochs for the Seq2Seq model and 30 epochs for REINFORCE) till convergence.
Most of the time was spent on RL training, which is over 3,633 minutes.
In constraints, when we tried to train CIPITR~\cite{saha2019complex}, the model required over 24 hours to complete one epoch of training while the max number of epochs is also set as 30.


Training with annotations would make the model learn to search in a relatively more accurate space, thus converging faster.
However, the limited availability of annotations remains a bottleneck for model training in many CQA tasks.
On the other hand, training without annotations but with denotations solely makes model convergence harder. 

We married the two ideas together: training with a small number of annotations and then the denotations.
First, we automatically produced pseudo-gold annotations for a small set (e.g., less than 1\% of the entire CQA training dataset) of questions. 
The pseudo-gold annotations were utilized to pre-train the model to constrain the search space.
After that, the model was further trained with only denotations.

\subsection{Results on CQA}
\label{cqa_result}
Table~\ref{tab2} summarizes the performance in F1 of the four models on the full test set of CQA.

It must be pointed out that \textbf{CIPITR}~\cite{saha2019complex} separately trained \emph{one single model for each of the seven question categories}. 
We denote the model learned in this way as \textbf{CIPITR-Sep}.
In testing, CIPITR-Sep obtained test results of each category by employing the corresponding tuned model~\cite{saha2019complex}. 
In practical use, when trying to solve a complex question more precisely, CIPITR-Sep has to first trigger a classifier to recognize the question category.
Only after acquiring the question categories information could CIPITR-Sep know which model to select to answer the question. 
If the number of question categories is increased, CIPITR-Sep needs to train more models, which will impede the system from generalizing to unseen instances. 
Besides, CIPITR also trained \emph{one single model over all categories of training examples} and used this single model to answer all questions. 
We denote this single model as \textbf{CIPITR-All}. 
Therefore, we separately present the performance of these two variations of CIPITR in Table~\ref{tab2}. 
On the other hand, we tuned NS-CQA on all categories of questions with one set of model parameters. 
Our model is designed to adapt appropriately to various categories of questions with one model, thus only needs to be trained once.

We also compared our full model, NS-CQA, with several model variants to understand the effect of our techniques presented in this work. 
Specifically, \textbf{Vanilla} is an imitation-learning model that was trained with pseudo-gold annotations. 
\textbf{PG} denotes the RL model that was optimized by the Policy Gradient algorithm based on the pre-trained model Vanilla. 
\textbf{NS-CQA} means the RL model that is equipped with all the techniques proposed in this work, notably the memory buffer and the reward bonus.

In Table~\ref{tab2}, several important observations can be made.

\begin{enumerate}
\item Over the entire test set, our full model NS-CQA achieves the best overall performance of 71.01\% and 80.80\% for macro and micro F1, respectively, outperforming all the baseline models.
The performance advantage on macro F1 over the four baselines is especially pronounced, by 51.56, 51.19, 23.31, and 8.62 percentage points over HRED+KVmem, CIPITR-All, CIPITR-Sep, and NSM respectively.
Also, NS-CQA improves over the micro F1 performance of HRED+KVmem, CIPITR-All, CIPITR-Sep, and NSM by 49.62\%, 49.28\%, 7.49\%, and 4.79\%. 
Moreover, our model achieves best or second-best in all the nine items being evaluated (the seven categories, plus overall macro F1, and overall micro F1).
The improvement is mainly due to the techniques presented in this work.
We introduce masking and copy mechanisms to reduce the search space effectively and carefully design a set of primitive actions to simplify the trials, therefore enable the model to efficiently find the optimal trials.
We also augment the RL model with a memory buffer, whereby the model could circumvent the spurious challenge, and remember the high-reward trials to re-sample them. 

\item Out of the seven categories, our full model NS-CQA achieves the best performance in four categories: Quantitative Reasoning, Verification (Boolean), Quantitative (Count), and Comparative (Count), and second best in the rest three. 
In the hardest categories, Quantitative (Count) and Comparative (Count), NS-CQA is substantially superior over the four baseline models, and outperforms our PG model. 
Since the length of the questions in the hardest categories is usually higher than in the other categories, it is always hard to find correct trials.
Under such circumstances, the memory buffer could make the model search in unknown space while keeping the previous high-reward trials in mind, which makes the model easier to train.
This is the main reason that NS-CQA performs the best in the hardest categories.

\item CIPITR-Sep achieves the best performance in two \emph{easy} categories, including the largest type, Simple Question. 
For the harder categories, it performs poorly compared to our model. 
Also, CIPITR-All, the single model that is trained over all categories of questions, performs much worse in all the categories than CIPITR-Sep, which learns a different model separately for each question category. 
For CIPITR-Sep, the results reported for each category are obtained from the model explicitly tuned for that category. 
A possible reason for CIPITR-All's significant performance degradation is that the model tends to forget the previously appeared high-reward trials when many infeasible trials are generated. 
Besides, the imbalanced classes of questions also deteriorates the performance of the model. 
Different from CIPITR, our model is designed to remember the high-reward trials when training. 

\item NSM and NS-CQA both produce competitive results.
The copy mechanism, masking method, and our carefully-defined primitive actions presented in this work were used in both models when we implemented them. 
By comparing the overall macro and micro F-1 score, it could be observed that NSM performed the best in all the four baseline models. 
This helps to validate the effectiveness of our proposed techniques.
However, NSM is worse than NS-CQA in all categories, especially in the harder ones. 
Since NSM records one promising trial for each question, it might be faced with the spurious problem. 
Different from NSM, we design a memory buffer for recording all successful trials to circumvent this problem. 
Also, NSM only considers the correctness of the predicted answers when measuring the reward, hence suffers from the sparse reward problem. 
Unlike NSM, our NS-CQA model augments the reward with proximity and novelty to mitigate this problem. 
These two factors make our model superior to the NSM model in all question categories.

\item Both of our model variants perform competitively. 
In Table~\ref{tab2}, it can be seen that the PG model, which was equipped with RL, performed better than the Vanilla model in five categories, but did not perform well in the two easy categories, Simple and Logical Reasoning. 
We analyzed the degeneration and found that for these two types of questions, usually, each question has only one correct sequence of actions.
When training with PG, some noisy spurious trials were introduced by beam search and thus degraded the model's performance. 
Our full model is better than the PG model in six of the seven categories and substantially improved performance in the Logical Reasoning category.
We compared the trials generated by the full model and the PG model, and found that many noisy trials are removed with the help of the memory buffer.
That is the main reason for the improvement in the Logical Reasoning category.
However, we also found that the full model performed worse than PG in Comparative Reasoning, which will be further investigated in the future.
\end{enumerate}

The above results demonstrate the effectiveness of our technique. 
It is worth noting that our model is trained on only 1\% of the training samples, whereas the baseline models use the entire training set. 
Besides, our approach uses one model to solve all questions, while CIPITR-Sep trains seven separate models to solve the seven categories of questions. 
Thus our model is virtually compared with seven individually training models used in CIPITR-Sep. 
However, our model still achieves best performance overall as well as in five of the seven categories.

\begin{table}[htb]
\caption{Performance comparison (measured in F1) of the four methods on the WebQuestionsSP test set. Best results is \textbf{bolded}.}\label{tab4}
\centering
\begin{tabular}{lr}
\toprule
{\textbf{Method}} & {\textbf{F1 measure}}\\ 
\midrule
CIPITR-All & 43.88\% \\
NSM & 70.61\% \\
PG & 70.72\% \\
NS-CQA & \textbf{72.04\%} \\
\bottomrule
\end{tabular}
\end{table}

\subsection{Results on WebQuestionsSP}
\label{sp_result}
Table~\ref{tab4} summarizes the performance in the F1 measure of the four models on the full test set of WebQuestionsSP.

Similar to the CQA dataset, CIPITR also divided questions into five categories, and then separately train one model for each category. 
However, since the category information is not provided in the WebQuestionsSP dataset, we did not classify the questions and trained one single model, CIPITR-All, for all the training samples by using its open-source code. 


From Table~\ref{tab4}, we can observe that NS-CQA can indeed learn the rules behind the multi-hop inference process directly from the distance supervision provided by the question-answer pairs. 
Without manually pre-defined constraints, our model could learn basic rules from the pseudo-gold annotations, and further complete the rules by employing RL. 

NS-CQA performed the best in the four models and significantly outperformed the CIPITR-All.
The main reason is that it is hard for CIPITR-All to learn one set of parameters that fits the different samples. 

Also, by introducing the masking and copy mechanism, NSM could achieve a performance competitive to our models PG and NS-CQA. 
By employing memory buffer, our NS-CQA model can alleviate the sparse reward problem and avert being trapped by spurious trials, which makes the model more robust, therefore achieving the best performance. 

Furthermore, NS-CQA achieves the best result on both the CQA and WebQuestionsSP datasets, which attests to the effectiveness and the generalizability of our method.

\subsection{Model Analysis}
To study how the different components influence the performance of our seq2seq model, i.e., Vanilla, we conduct an ablation experiment as follows. 
Each of the main components: attention, copy mechanism, and masking method, is removed one at a time from the full seq2seq model to study how its removal affects model performance. 
We also study the effect of smaller training samples on performance, by using 1K and 2K samples for training, instead of 10K used in the full model. 

Table~\ref{tab3} summarizes performance degradation on the CQA test set, where the Vanilla model achieves a macro F1 score of 62.15\%. 
It can be seen that the removal of masking produces the largest drop in performance, of 37.10\%. 
Masking method significantly decreases the search space by replacing all the entity, relation and type names with wildcard tokens. 
This result demonstrates that although a simple approach, masking proves to be valuable for the CQA task. 

\begin{table}[htb]
\caption{Ablation study on the CQA test set, showing the macro F1 score drop by removing each main component, or by learning from a subset of the training set. The Vanilla model has macro F1 of 62.15\% as shown in Table~\ref{tab2}.}\label{tab3}
\centering
\begin{tabular}{lr}
\toprule
{\textbf{Feature}} & {\textbf{Macro F1}}\\ 
\midrule
Vanilla & 62.15\%\\
Masking & -37.10\% \\
Copy & -11.52\% \\
Attention & -4.30\% \\
\midrule
1,000-training & -5.78\% \\ 
2,000-training & -4.09\% \\
\bottomrule
\end{tabular}
\end{table}

When the copy mechanism is removed, performance decreases by 11.52\%. 
This is consistent with our expectation since masking has already considerably decreased the search space, the improvements that copy mechanism could makes is relatively limited. 
Lastly, when training on fewer samples labeled with pseudo-gold actions, the model under-fits. 

When training on even smaller datasets, the performance degradation is not as severe as we expected. 
With as a training set as small as 1,000, our model is able to generalize well, only suffering a 5.78\% drop on a test set of 15.6K. 
With a training set of 2,000 samples, our model suffers a modest 4.09\% drop in performance. 
This study further demonstrates the robustness and generalizability of our model. 

\subsection{Sample Size Analysis}
In this subsection, we analyze the effect of training samples of different sizes on our PG module. 
Since the WebQuestionsSP consists of a limited number of questions, it is hard to conduct the sample size analysis on it.
Instead, we trained our model by using different CQA subsets to make a comparison.
Specifically, given the same pre-trained model, we train the NS-CQA model on 0.2\%, 0.4\%, 0.6\%, 0.8\%, 1.0\%, and 1.2\% of total 944K training samples. 
Note that the evaluation results of the full model presented in Section~\ref{cqa_result} are obtained from 1.0\% of training data and the entire test set (i.e., 156K). 
For experiments described in this subsection, evaluation is performed on a subset of the full test set that is 10\% of its size (i.e., 15.6K). 
Training of the REINFORCE model is stopped at 30 epochs, which is when all models have been observed to converge. 

\begin{figure}[htb]
\centering
\includegraphics[width=.5\textwidth,trim={0 2.9cm 0 3.2cm},clip]{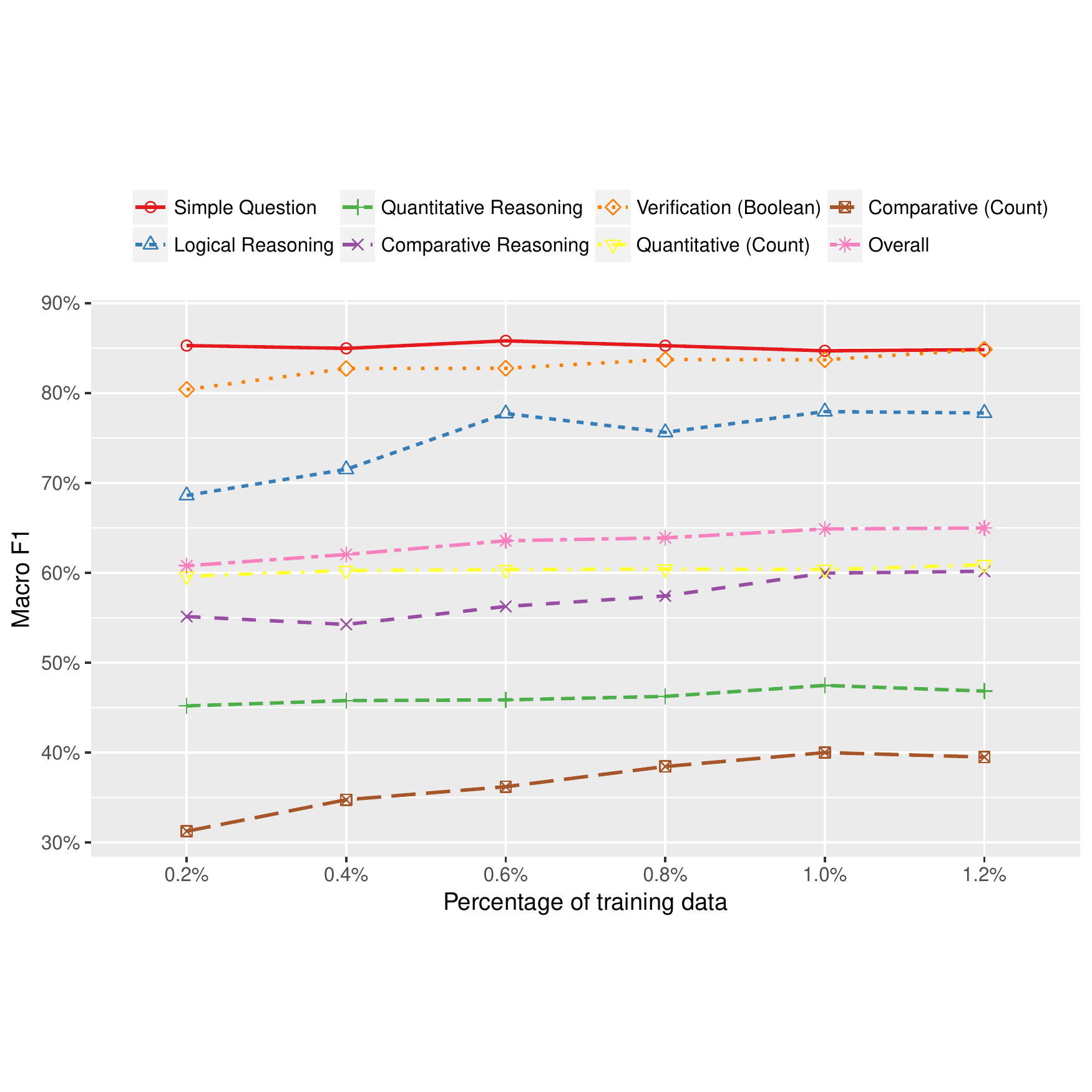}
\caption{Changes in macro F1 values with varying percentages of training data for the PG module.}\label{fig:f1_trend}
\end{figure}

We first study the effect on model performance. 
Figure~\ref{fig:f1_trend} plots the macro F1 values of the seven categories of questions as well as the overall performance. 
With the increase in training data size, a general upward trend in macro F1 values can be observed, with the category Simple Question being the exception. 
For the overall test set, we can observe that the macro F1 value plateaus at 1.0\% and does not increase when training data is expanded to 1.2\%.

For Simple Question, the output actions are relatively the `simplest'.
In most cases, one `Select' action is needed to solve a question. 
Consequently, with the help of methods to decrease search space and better use data, after pre-training, the model overfits on the Simple Question type rapidly. 
Therefore the performance of answering Simple Question fluctuates with the change of training sample size. 

On the other hand, for the other categories, it could be found that the model can use data efficiently and obtain the best result by training on only 1\% samples. 

\begin{figure}[htb]
\centering
\includegraphics[width=.5\textwidth,trim={0 5.3cm 0 5.3cm},clip]{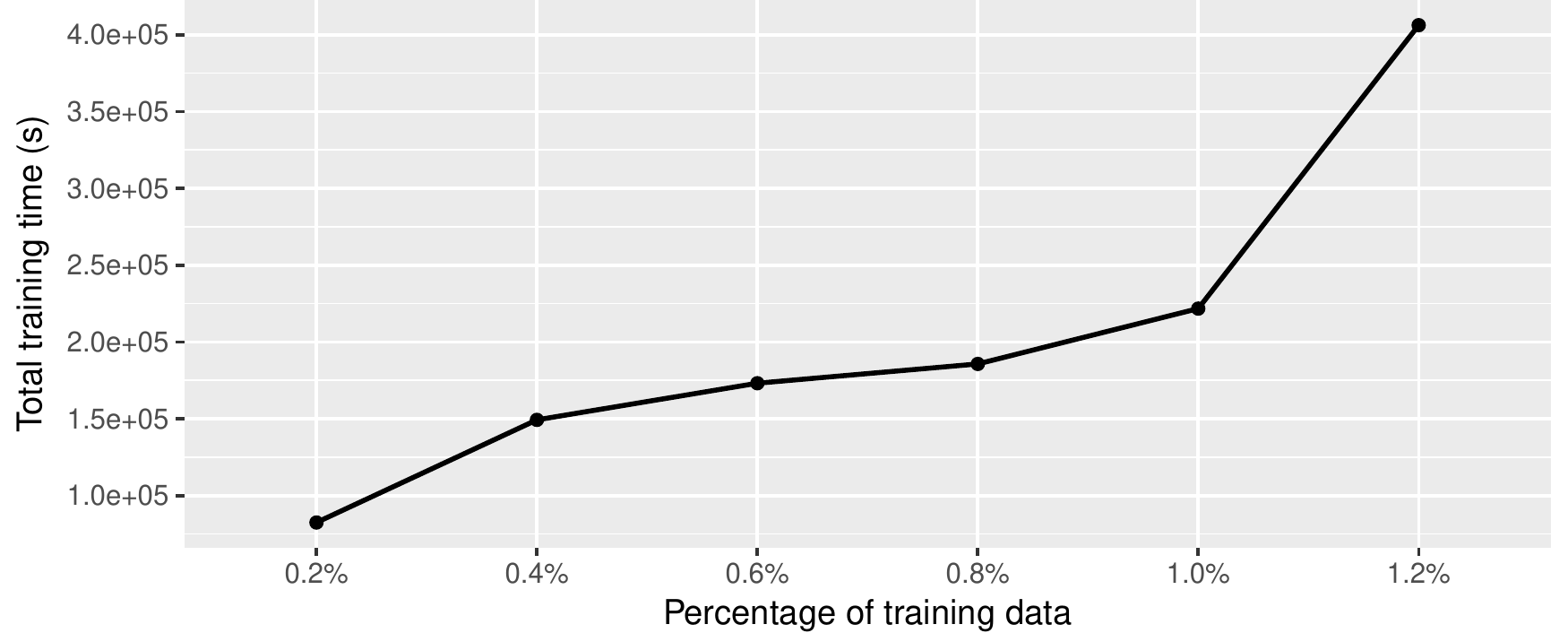}
\caption{Changes in training time (in seconds) with varying percentages of training data.}\label{fig:time_trend}
\end{figure}

More samples might help the model on some question categories, but more training time is consumed. 
The training time of the REINFORCE module is plotted in Figure~\ref{fig:time_trend}. 
As can be seen, there is a significant increase in training time when training data increases from 1.0\% to 1.2\%. 
Together with the trend of the macro F1 value, as shown in Figure~\ref{fig:f1_trend}, we can empirically determine the best trade-off between model performance and training efficiency at 1.0\%.

\section{Qualitative Analysis}\label{sec:qual_ana}
In this section, we analyze the quality of our NS-CQA model in more detail. We first present some success cases where NS-CQA can predict the action sequence that produces correct answers. A detailed analysis of typical errors is then performed, which sheds light on the areas that can be further investigated. 

\subsection{Sample Cases}

In Table~\ref{tab:cases} below, we present some example questions from different categories that our model NS-CQA can correctly predict action sequences. 

We can inspect the complexity of the CQA problem from these instances.
As is demonstrated in the table, Simple and Logical questions are simplest to answer since commonly, only 2-3 actions are needed.
The following four categories, i.e., Quantitative, Comparative, Verification, and Quantitative Count, are relatively more difficult types with around 3-4 operations.
For instance, the Verification question ``Is Alda Pereira-Lemaitre a citizen of France and Emmelsb\"{u}ll-Horsb\"{u}ll?'' has an answer ``YES and NO respectively''. 
Answering this question involves selecting all countries which Alda Pereira-Lemaitre is a citizen of, and verifying whether France and Emmelsb\"{u}ll-Horsb\"{u}ll is in this set respectively. 
The last type, Comparative Count, is the most complex for questions of which will be transformed into more than five actions. 

Evident from the Quantitative Count and the Comparative Count questions in the last two rows of the table, answering CQA questions involve discrete actions. 
In the case of the question ``How many assemblies or courts have control over the jurisdiction of the Free Hanseatic City of Bremen?'', the set operation Union is required. 
In the case of the question ``How many art genres express more number of humen or concepts than floral painting?'', numerical operations (GreaterThan and Count) are required. 

These example questions attest to the challenging nature of the CQA dataset and the capability of our NS-CQA model.

\begin{table*}[htb]
\centering
\caption{Examples of action sequences correctly predicted by NS-CQA for different types of questions.}
\label{tab:cases}
{\small
\begin{tabular}{|p{1.5cm}|p{3.5cm}|p{3.5cm}|l|p{3cm}|}
\hline 
\textbf{Q.\ type} & \textbf{Question} & \textbf{KB artifacts} & \textbf{Action sequence} & \textbf{Answer}\\ \hline
Simple & Which administrative territory is Danilo Ribeiro an inhabitant of? & \begin{tabular}[t]{@{}l@{}l@{}}\textbf{E1:} Danilo Ribeiro\\\textbf{R1:} country of citizenship\\\textbf{T1:} administrative territory\end{tabular}& \begin{tabular}[t]{@{}l@{}}Select(E1, R1, T1)\\EOQ\end{tabular} & Brazil\\ \hline
Logical & Which administrative territories are twin towns of London but not Bern? & \begin{tabular}[t]{@{}l@{}l@{}l@{}}\textbf{E1:} London\\\textbf{E2:} Bern\\\textbf{R1:} twinned adm.\ body\\\textbf{T1:} administrative territory\end{tabular} & \begin{tabular}[t]{@{}l@{}l@{}}Select(E1, R1, T1)\\Diff(E2, R1, T1)\\EOQ\end{tabular} & Sylhet, Tokyo, Podgorica, Phnom Penh, Delhi, Los Angeles, Sofia, New Delhi, \ldots\\ \hline
Quantitative & Which sports teams have min number of stadia or architectural structures as their home venue? & \begin{tabular}[t]{@{}l@{}l@{}l@{}}\textbf{R1:} home venue\\\textbf{T1:} sports team\\\textbf{T2:} stadium\\\textbf{T3:} architectural structure\end{tabular} & \begin{tabular}[t]{@{}l@{}l@{}l@{}}SelectAll(T1, R1, T2)\\SelectAll(T1, R1, T3)\\ArgMin()\\EOQ\end{tabular}& Detroit Tigers, Drbak-Frogn IL, Club Sport Emelec, Chunichi Dragons, \ldots\\ \hline
Comparative & Which buildings are a part of lesser number of architectural structures and universities than Midtown Tower? & \begin{tabular}[t]{@{}l@{}l@{}l@{}l@{}}\textbf{E1:} Midtown Tower\\\textbf{R1:} part of\\\textbf{T1:} building\\\textbf{T2:} architectural structure\\\textbf{T3:} university\end{tabular} & \begin{tabular}[t]{@{}l@{}l@{}l@{}}SelectAll(T1, R1, T2)\\SelectAll(T1, R1, T3)\\LessThan(E1)\\EOQ\end{tabular} & Amsterdam Centraal, Hospital de Sant Pau, Budapest Western Railway Terminal, El Castillo, \ldots\\ \hline
Verification & Is Alda Pereira-Lemaitre a citizen of France and Emmelsb\"{u}ll-Horsb\"{u}ll? & \begin{tabular}[t]{@{}l@{}l@{}l@{}l@{}}\textbf{E1:} Alda Pereira-Lemaitre\\\textbf{E2:} France\\\textbf{E3:} Emmelsb\"{u}ll-Horsb\"{u}ll\\\textbf{R1:} country of citizenship\\\textbf{T1:} administrative territory\end{tabular} & \begin{tabular}[t]{@{}l@{}l@{}l@{}}Select(E1, R1, T1)\\Bool(E2)\\Bool(E3)\\EOQ\end{tabular}& YES and NO respectively\\ \hline
Quantitative Count & How many assemblies or courts have control over the jurisdiction of Free Hanseatic City of Bremen? & \begin{tabular}[t]{@{}l@{}l@{}l@{}}\textbf{E1:} Bremen\\\textbf{R1:} applies to jurisdiction\\\textbf{T1:} deliberative assembly\\\textbf{T2:} court\end{tabular} & \begin{tabular}[t]{@{}l@{}l@{}l@{}}Select(E1, R1, T1)\\Union(E1, R1, T2)\\Count()\\EOQ\end{tabular}& 2\\ \hline
Comparative Count & How many art genres express more number of humen or concepts than floral painting? & \begin{tabular}[t]{@{}l@{}l@{}l@{}l@{}}\textbf{E1:} floral painting\\\textbf{R1:} depicts\\\textbf{T1:} art genre\\\textbf{T2:} human \\\textbf{T3:} concept\end{tabular} & \begin{tabular}[t]{@{}l@{}l@{}l@{}l@{}}SelectAll(T1, R1, T2)\\SelectAll(T1, R1, T3)\\GreaterThan(E1)\\Count()\\EOQ\end{tabular}& 8\\ \hline
\end{tabular}
}
\end{table*}

\subsection{Error Analysis}
To analyze the limitations of our NS-CQA model, 200 samples in each category that produce incorrect answers are randomly selected from the test dataset. 
In summary, a large number of errors can be categorised into one of the following five classes. 

\subsubsection{Linking Problem}
Since different entities/types might have the same surface name, in addition to literal similarity, the embedding of the description of entities/types and the embedding of question is employed to compute semantic similarity in our approach. 
When mapping the predicates to queries, a state-of-the-art convolutional sequence to sequence (Seq2Seq) learning model~\cite{gehring2017convolutional} implemented in fairSeq~\cite{ott2019fairseq} is used. 
Even so, some linking problems remain.

\paragraph{Example:} ``Where are around the same number of geographic locations located on as Big Salmon Range?''. 

When our model is answering the above question, the relation `located on street' is wrongly linked to the question instead of the correct relation `located on terrain feature'. 
This type of errors can be addressed by learning better semantic meaning of the entities/types/relations from the context in the knowledge graph. 

\subsubsection{Infeasible Action}
NS-CQA occasionally produces meaningless and repetitive actions which are semantically incorrect. 
For instance, some actions are predicted to union the same set, which is reluctant. 
In some cases, two repeated `Count' actions are predicted. 

\paragraph{Example:} ``What social groups had Canada and Austria as their member?''

When our model is solving the above question, it predicts the following action sequence: 

{\small
\begin{quote}
Select (Canada, member of, social group)\\
Bool (Austria)\\
EOQ
\end{quote}} 

The operator `Bool' is invalid since in this question the expected answer type is entities but not Boolean values. 
Semantic-based constraints could be employed to make the model produce feasible actions. 

\subsubsection{Spurious Problem}
In our approach, the pseudo-gold action sequences are generated by a BFS algorithm. 
Therefore corresponding to each question, multiple possible sequences may evaluate to the same expected results. 
Among these sequences, there might be some spurious action sequences. 
When training the model with such action sequences, the model may be misled and produce incorrect actions. 

\paragraph{Example:} ``Which musical ensembles were formed at Belfast?''

Our model transforms the above question into the following action sequence: 

{\small
\begin{quote}
Select(Belfast, location, musical ensemble)\\
Inter(Belfast, location, musical ensemble)\\
EOQ 
\end{quote}
}
The second action `Inter' is unnecessary to this question. 
Rule-based constraints could be incorporated to restrict the search process to meaningful actions. 

\subsubsection{Order of Arguments}
To decide the order of the entities/types in the actions is a difficult problem. 
For actions `Select', `Inter', `Diff', and `Union', the order of the arguments is decided by the following rule: the first argument in a triple pattern is related to the entity, and the last argument is associated with the type. 
In most cases, a sequence of entities/types in actions follows the order they appear in the question. 
Though our model is also trained to handle the situation that the sequence of entities/types does not appear in the same order, in some cases, the model is confused about which order to follow.

\paragraph{Example:} ``Is Bernhard II, Duke of Saxe-Jena a child of William, Duke of Saxe-Weimar?''

Our model transforms the above question into the following action sequence: 

{\small
\begin{quote}
Select (Bernhard II. Duke of Saxe-Jena, child, common name)\\
Bool (William. Duke of Saxe-Weimar)\\
EOQ 
\end{quote}
}

However, the correct action sequence should be the following. As can be seen, the order of the two entities is wrong in the predicted sequence. 

{\small
\begin{quote}
Select (William. Duke of Saxe-Weimar, child, common name)\\
Bool (Bernhard II. Duke of Saxe-Jena)\\
EOQ
\end{quote}
}

In future work, we will investigate whether incorporating more positional information can help alleviate this problem.  

\subsubsection{Approximation-related Problem}
The action `Almost' is used to find the set of entities whose number is approximately the same as a given value, and such operation appears in the following four categories of questions: Quantitative Reasoning, Quantitative Count, Comparative Reasoning, and Comparative Count. 
The questions involving the `Almost' action account for 4\% of the total test dataset. 
When solving such questions, the range of the approximate interval is naturally vague. 
We define the following ad-hoc rule to address this vagueness: suppose we are required to find the value around $N$, when $N$ is no larger than 5, the interval is $[N-1, N+1]$; when $N$ is more significant than 5, the range is $[N-5, N+5]$.  
In some cases, this rule works, but in others not.

\paragraph{Example:} ``Which political territories have diplomatic relations with approximately 14 administrative territories?''

The following action sequence could be produced to solve such questions: 

{\small
\begin{quote}
SelectAll (political territorial entity, diplomatic relation, administrative territorial entity)\\
Almost (14)\\
EOQ 
\end{quote}
}

Following our rule, the approximate interval here should be [9, 19]. 
However, the correct answer (political territorial entities) may have several administrative territories outside this range. 
Thus, the unfixed approximate interval may impair the performance of our model. 
We can manually tweak the rule of deciding the approximate interval.
However, we emphasize that our model aims to show a robust framework to solve complex questions, but not to guess rules for approximation. 

\section{Conclusion}
\label{sec:conclusion}
Answering complex questions on KBs is a challenging problem as it requires a model to perform discrete operations over KBs.
State-of-the-art techniques combine neural networks and symbolic execution to address this problem. 
While practical, the challenges of these techniques reside in data-inefficiency, reward sparsity, and ample search space.

In this paper, we propose a data-efficient neural-symbolic model for complex KBQA that combines simple yet effective techniques, addressing some of the above deficiencies.

Firstly, we augment the model with a memory buffer.
When the memory buffer maintains the generated successful trials for each training question, it will guide the model to replay and re-sample the promising trials more frequently, thus mitigating the data-inefficiency problem.

Secondly, by comparing the generated trials with the trials stored in the memory, we assign a bonus to the reward, which is the combination of proximity and novelty.
Also, we propose an adaptive reward function.
The reward bonus and the adaptive reward reshape the sparse reward into dense feedback that can efficiently guide policy optimization.
Employing the curriculum-learning scheme, we gradually increase the proportion of proximity while decreasing the weight of novelty.
By doing this, we encourage the model to find new trials while remembering the past successful trials.

Thirdly, we incorporate the copy and masking mechanisms in the model, and carefully design a set of primitive actions, to drastically reduce the size of the decoder output vocabulary by orders of magnitude.
This significant reduction improves not only training efficiency but also model generalizability.
Also, our actions free the model from the need to maintain complex intermediate memory modules, thus simplifies network design.

We conduct experiments on two challenging datasets on complex question answering. 
In comparison with three state-of-the-art techniques, our model achieves the best performance and significantly outperforming them in both the datasets. 

\section*{Acknowledgements} 
This work was partially supported by the National Key Research and Development Program of China under grants (2018YFC0830200), the Natural Science Foundation of China grants (U1736204, 61602259), the Judicial Big Data Research Centre, School of Law at Southeast University, and the project no. 31511120201 and 31510040201.

\bibliographystyle{elsarticle-num-names}
\bibliography{bibfile}

\end{document}